\documentclass[10pt,twocolumn,letterpaper]{article}

\usepackage{iccv}
\usepackage{times}
\usepackage{epsfig}
\usepackage{graphicx}
\usepackage{amsmath}
\usepackage{amssymb}
\usepackage{breqn}
\usepackage{booktabs,siunitx} 
\usepackage{mathtools}

\DeclarePairedDelimiter\ceil{\lceil}{\rceil}
\DeclarePairedDelimiter\floor{\lfloor}{\rfloor}
\usepackage{pifont}
\newcommand{\cmark}{\ding{51}}%
\usepackage{arydshln}
\newcommand{\vect}[1]{\mathbf{#1}}

\usepackage{subfig}

\usepackage[pagebackref=true,breaklinks=true,letterpaper=true,colorlinks,bookmarks=false]{hyperref}

\iccvfinalcopy 

\def\iccvPaperID{1740} 

\ificcvfinal\fi

\begin{document}

\title{Hyperspectral Image Super-Resolution with Spectral Mixup and Heterogeneous Datasets}

\author{Ke Li$^1$\hspace{6mm} Dengxin Dai$^1$ \hspace{6mm} Ender Konukoglu$^1$ \hspace{6mm} Luc Van Gool$^{1,2}$ \\ 
 $^1$CVL, ETH Zurich,   $^2$PSI, KU Leuven\\
{\tt\small {\{ke.li,dai,ender.konukoglu,vangool\}@vision.ee.ethz.ch}}
}

\maketitle
\begin{abstract}
   This work studies Hyperspectral image (HSI) super-resolution (SR). HSI SR is characterized by high-dimensional data and a limited amount of training examples. This exacerbates the undesirable behaviors of neural networks such as memorization and sensitivity to out-of-distribution samples. This work addresses these issues with three contributions. First, we observe that HSI SR and RGB image SR are correlated and develop a novel multi-tasking network to train them jointly so that the auxiliary task RGB image SR can provide additional supervision. Second, we propose a simple, yet effective data augmentation routine, termed Spectral Mixup, to construct effective virtual training samples to enlarge the training set. Finally, we extend the network to a semi-supervised setting so that it can learn from datasets containing only low-resolution HSIs. With these contributions, our method is able to learn from heterogeneous datasets and lift the requirement for having a large amount of HD HSI training samples. Extensive experiments on four standard datasets show that our method outperforms existing methods significantly and underpin the relevance of our contributions. Code has been made available at \url{https://github.com/kli8996/HSISR}.
\end{abstract}

\section{Introduction}

Hyperspectral imaging acquires images across many intervals of the electromagnetic spectrum. It has been applied to numerous areas such as medical diagnosis~\cite{medical:review:14}, food quality and safety control~\cite{food:review:07}, remote sensing~\cite{remote:sensing:09} and object detection~\cite{hyspectraobjdetection}. All these applications benefit from analyzing the spectral information coming with HSIs. One obstacle in the way of further unleashing this potential is data acquisition.  Acquiring HSIs of high spatial and high spectral resolution at a high frame rate is still a grand challenge. There is still no camera to achieve these three goals at the same time. Cameras for a compromise setting -- high spectral but low spatial resolution -- are quite common by now, though still expensive. As a result, increasing efforts have been made to advance HSI super-resolution (SR).

While numerous deep learning methods have been developed for improving the resolution of RGB images (RGBIs), the topic of HSI SR has received little attention. One of the main reasons is the lack of large-scale HSI datasets for high-resolution (HR) HSIs. As known, supervised deep learning methods need an enormous amount of training data. This situation, unfortunately, will not be improved in the foreseeable future due to the challenges hyperspectral imaging faces. In this work, we choose a different route and propose to learn from multiple heterogeneous datasets and also from virtual examples. We find that while it is difficult to collect HR HSIs, it is relatively easy to collect only LR HSIs and it is very easy to collect HR RGB images. It is thus appealing to have a HSI SR method which can learn from these heterogeneous sources. Our method is designed for this aim.

Although the data distribution is not the same between RGBIs and HSIs, the two SR tasks do share some common goals in integrating information from neighboring spatial regions and neighboring spectral bands during the learning. We embrace this observation and formulate both tasks into the same learning framework such that the parameter distribution induced by the RGBI SR task can serve as an effective regularization for our HSI SR task. The challenge lies in the difference in spectral band numbers, e.g. three in RGBIs vs. e.g. 31 or 128 in HSIs. To tackle this and to reduce the computational complexity, we propose a universal group convolutional neural network that can accommodate different spectral groups. 

We further propose a data augmentation routine, termed \emph{Spectral Mixup}, to create effective `virtual' training examples. Data augmentation is a strategy to create virtual samples by modifying the original samples. Data augmentation is known to increase the generalizability of learning methods. Common methods for classification tasks include reflections, rotations, cropping, and color jittering. They assume that examples obtained by those operations share the same class with the original example and that can hardly be applied to our regression task. 
In this work, we propose \emph{Spectral Mixup} to create virtual samples using convex combinations of spectral bands of the same image for our task HSI SR. \emph{Spectral Mixup} favors functions that preserve simple linear behavior in-between spectral bands and greatly avoids data over-fitting.   

While the aforementioned contributions can yield state-of-the-art performance already, we extend the method further to learn from unlabeled datasets as well. 
Semi-supervised learning (SSL) exploits unlabeled data to reduce over-fitting to the limited amount of labeled data \cite{Dai_2013_ICCV,temporal:ensembling,mean:teacher:nips17,noisy:student:20,hoyer2020three}. While good progress has been achieved, the strategies are mainly designed for recognition tasks. Their applicability to a low-level dense regression task such as HSI SR has yet to be verified. In this work, we again leverage the success of RGBI SR and propose a cross-model consistency that favors functions giving consistent outputs between super-resolved RGBIs and super-resolved HSIs. Basically, we convert LR HSIs into LR RGB images and pass those through the RGBI SR network. In the meanwhile, we pass the LR HSIs through our HSI SR network to get the super-resolved HSIs and convert them to RGBIs with a standard camera response function. We enforce the consistency between the two versions of super-resolved RGBIs. This way, supervision is transferred from the better-trained RGB SR network to our HSI SR network via a second route.

To summarize, this work makes three contributions: 1) a multi-tasking HSI SR method to learn together with an auxiliary RGBI SR task;  2) a simple, yet effective data augmentation method \emph{Spectral Mixup}; and 3) A SSL method to learn also from `unlabeled' LR HSIs. With these contributions, our method sets the new state of the art.

\section{Related Work}


\noindent
\textbf{Hyperspectral Image Super-Resolution}.
HSI SR can be grouped into three categories according to their settings: 1) HSI SR from only RGBIs; 2) Single HSI SR from LR HSIs; and 3) HSI SR from both HR RGBIs and LR HSIs of the same scene. Our method belongs to the second group. 

HSR SR from only RGBIs is a highly ill-posed problem. However, it has gained great traction in recent years due to its simple setup and  the well-organized workshop challenges~\cite{Arad_2020_CVPR_Workshops}. Similar to other computer vision topics, the trend has shifted from `conventional' methods such as radial basis functions~\cite{Training:Spectral:rgb:14} and sparse coding~\cite{arad_and_ben_shahar_2016_ECCV} to deep neural networks~\cite{Galliani2017LearnedSS,HSCNN+,Arad_2020_CVPR_Workshops}. This trend highlights the need for bigger training datasets.


Single image SR aims to model the relationship between the LR images and HR ones by learning from a collection of examples consisting of pairs of HR images and LR images. Single RGBI SR has achieved remarkable results in the last years. Since the first work of using neural networks for the task~\cite{sr:eccv14}, progress has been made in making networks deeper and the connections denser~\cite{very:deep:SR:16,residual:dense:sr:cvpr18}, using feature pyramids~\cite{deep:laplacian:cvpr17}, employing GAN losses~\cite{Ledig_2017_CVPR}, and modeling real-world degradation effects~\cite{guo2020closed}.  
As to single HSI SR, there has been great early work~\cite{HSI:SR:05,HIS:SR:2011} as well. However, that is also surpassed by deep learning methods. For instance, Yuan \etal~\cite{HSI:SR:TL:17} trained a single-band SR method on natural image datasets, and applied it to HSIs in a band-wise manner to explore spatial information. The spectral information is explored via matrix factorization afterwards. In order to explore both spatial and spectral correlation at the same time, methods based on 3D Convolutional Networks~\cite{3d:net:HSI:sr:17,2d:3d:net:HSI:SR:20} have been developed. Although 3D CNNs sound like a perfect solution, the computational complexity is very high. To alleviate this, Grouped Convolutions (GCs) with shared parameters have been recently used in~\cite{HSI:SR:grouped:recursivenet:18,spatial:spectral:prior:20}. The backbone network of our method is also based on GCs.


Fusion-based methods use HR RGBIs of the same scene as references to improve the spatial resolution of the LR HSIs~\cite{unmixing:survey:12,HSISR:fusion:survey:17}. This stream of methods have received more research attention than the former two. Many learning techniques have been applied to this data fusion task including Bayesian inference~\cite{bayesian:HSISR:15,HSISR:beta:eccv16,Zhang_2020_CVPR}, matrix factorization~\cite{HSI:unmixing:iccv15,HSI:SR:tensor:17}, sparse representation~\cite{HSISR:sparse:eccv14,HSI:sparse:coding:tip16}, and deep neural networks~\cite{unsueprvised:fusion:cvpr18,fusion:net:cvpr19}. The common goal of these methods is to learn to propagate the detailed information in the HR RGBIs to the target HSIs and fuse them with the fundamental spectral information from LR HSIs. Despite the plethora of fusion algorithms developed, they all assume that the LR HSIs and the HR RGBIs are very well co-registered~\cite{spatial:spectral:prior:20}. This data registration is a challenge on its own and registration errors will lead to degraded SR results~\cite{data:fusion:15,fusion:20}.







 \begin{figure*}[!tb]
    \centering
    \includegraphics[width=.95\textwidth]{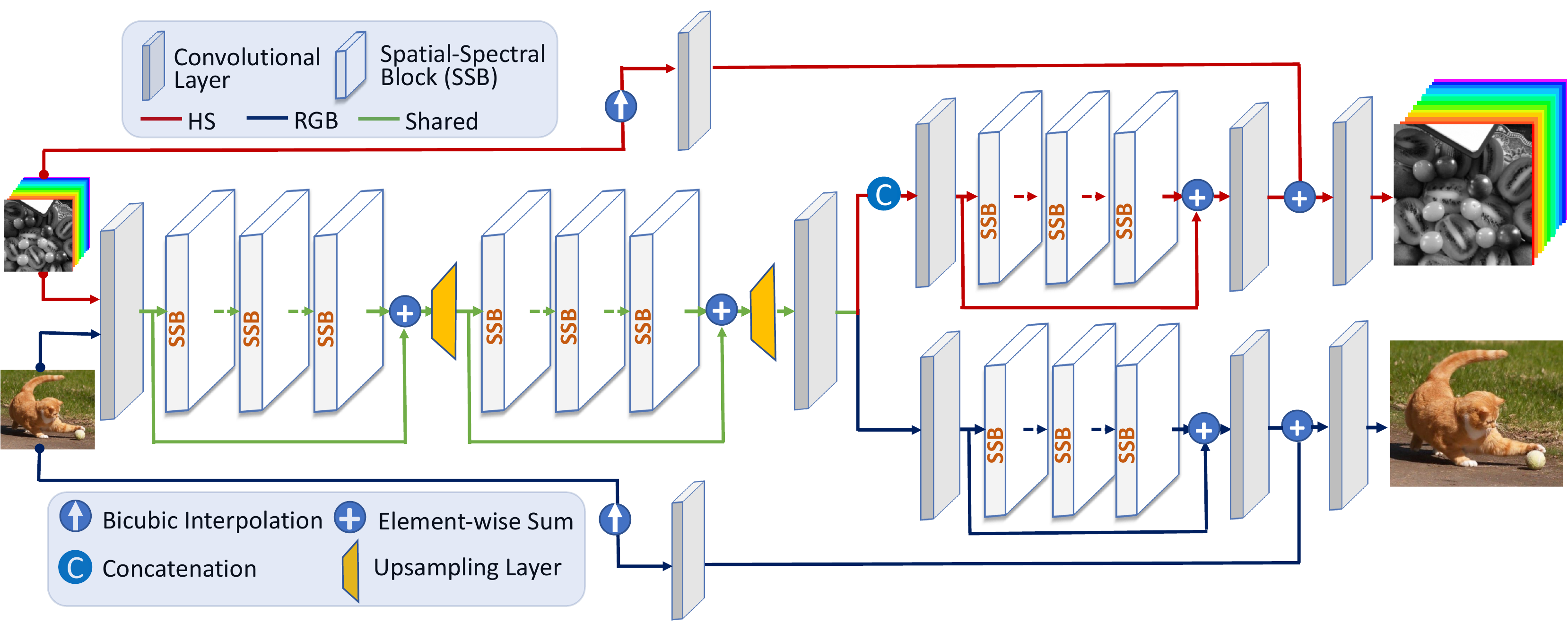}
    \caption{The architecture of our network consisting of a shared encoder and two specific decoders for the two SR tasks.}
    \label{fig:network}
\end{figure*}

\noindent 
\textbf{Learning with Auxiliary Tasks}. 
It is quite a common practice to borrow additional supervision from related auxiliary tasks, when there is insufficient data to learn a task. The common strategy is to learn all the tasks together so that the auxiliary tasks can regularize the optimization. There are normally two assumptions: (1) we only care about the performance of the main task and (2) the supervision for the auxiliary tasks is easier to obtain than that of the main task. Previous work has employed various kinds of self-supervised methods as auxiliary tasks for the main supervised task in a semi-supervised setting~\cite{ssl:gan:14,s4l:iccv19,auxiliary:task:nips20}. For instance, generative approaches have been explored in~\cite{ssl:gan:14} and predicting the orientation of image patches is used in~\cite{s4l:iccv19}. 
Another related setting is multi-task learning (MTL)~\cite{mtl:survey:20}. In MTL, the goal is to reach high performance on multiple tasks simultaneously, so all tasks are main tasks and all tasks are auxiliary tasks. While the goal is different, many strategies in MTL such as parameter sharing~\cite{semanticbinaural}, task consistency~\cite{zamir2020consistency}, and loss balance~\cite{mtl:loss:cvpr18} are useful for learning with auxiliary tasks. 


 

\section{Approach}
\label{sec:approach}
HSIs provide tens of narrow bands, so processing all the bands together is time-consuming and requires very large datasets in order to avoid over-fitting. In this work, we follow~\cite{spatial:spectral:prior:20} and use a grouping strategy to divide input HSIs into overlapping groups of bands. This way, the spectral correlation among neighboring bands can be effectively exploited without increasing the parameters of the model. Another major advantage of using a grouping strategy is that it offers the possibility to train our auxiliary task RGBI SR along with our main task HSI SR within the same network. Without using the grouping strategy, the difference in the number of bands is very large between the two tasks. In this work, we assume that the relationships between low/high-resolution HSIs and low/high-resolution RGBIs are correlated, so they should be trained together so that RGBI SR can provide additional supervision for HSI SR. This way, the HSI SR method can enjoy training samples of a much more diverse set of scenes especially those that cannot be captured well by current hyperspectral imaging devices such as moving objects.

\subsection{HSI SR with an Auxiliary RGBI SR Task}
\label{sec:basic:network} 

Given two SR tasks $\mathcal{T}_{\text{HS}}$ and $\mathcal{T}_{\text{RGB}}$, we aim to help improve the learning of a model for $\mathcal{T}_{\text{HS}}$ by using the knowledge contained in $\mathcal{T}_{\text{RGB}}$. In the supervised setting, each task is accompanied by a training dataset consisting of $N$ training samples, i.e., $\mathcal{D}_{\text{HS}}=\{\vect{x}_{\text{HS}}^i, \vect{X}_{\text{HS}}^i\}_{i=1}^{N_{\text{HS}}}$ and $\mathcal{D}_{\text{RGB}}=\{\vect{x}_{\text{RGB}}^i, \vect{X}_{\text{RGB}}^i\}_{i=1}^{N_{\text{RGB}}}$, where $\vect{x}_{\text{HS}} \in \mathbb{R}^{h_1\times w_1 \times C}$, $\vect{X}_{\text{HS}} \in \mathbb{R}^{H_1\times W_1 \times C}$,  $\vect{x}_{\text{RGB}} \in \mathbb{R}^{h_2\times w_2 \times Z}$, and $\vect{X}_{\text{RGB}} \in \mathbb{R}^{H_2\times W_2 \times Z}$. 
We denote low-resolution (LR) images by $\vect{x}$, high-resolution (HR) images by $\vect{X}$, the number of bands of HSIs by $C$, the number of bands in RGB images by Z (3 here), and the size of the images by $h$, $w$, $H$ and $W$. Given a scaling factor $\tau$, we have $H_{i}=\tau h_{i}$ and $W_{i}=\tau w_{i}$ for both tasks. 

The goal is to train a neural network $\Phi_{\text{HS}}$ to predict the HR HSI for a given LR HSI:  $\vect{X}_{\text{HS}} = \Phi_{\text{HS}}(\vect{x}_{\text{HS}})$. Different from previous methods, which have a single network for the whole task, our method consists of three blocks: an encoder which is shared by the two SR tasks, and two task-specific decoders to output the final outputs. More specifically, $\Phi_{\text{HS}}=(\Phi^{\text{En}}, \Phi^{\text{De}}_{\text{HS}})$ and  $\Phi_{\text{RGB}}=(\Phi^{\text{En}}, \Phi^{\text{De}}_{\text{RGB}})$. 
The general architecture is shown in Fig~\ref{fig:network}. 

In order to share the same encoder between the two SR tasks, we divide the $C$ input channels of $\vect{x}_{\text{HS}}$ into groups of $M$ bands. For HSI SR, the encoder network $\Phi^{\text{En}}$ takes $M$ channels as input and generates $M$ channels as output. The outputs of all the groups of $\vect{x}_{\text{HS}}$ are then concatenated according to their original spectral band position to assemble a new HSI $\bar{\vect{X}}_{\text{HS}} \in \mathbb{R}^{H_1\times W_1\times C}$. The neighboring groups of $\vect{x}_{\text{HS}}$ can have overlaps and we average the results of the overlapping areas when assembling $\bar{\vect{X}}_{\text{HS}}$. There are two upsampling layers to upscale the size of the input to the desired size in a progressive manner. This progressive upsampling has proven useful for both RGBI SR~\cite{deep:laplacian:cvpr17} and HSI SR~\cite{spatial:spectral:prior:20}.  The reconstructed $\bar{\vect{X}}_{\text{HS}}$ is then fed into the decoder network $\Phi^{\text{De}}_{\text{HS}}$ as a whole to generate the final output  $\hat{\vect{X}}_{\text{HS}}$, which is then compared to the ground truth $\vect{X}_{\text{HS}}$ to compute the loss. $\Phi^{\text{De}}_{\text{HS}}$ takes all the bands directly to learn long-range spectral correlations beyond individual groups to  refine the results.

For RGBI SR, we first increase the number of bands of $\vect{x}_{\text{RGB}}$ from $Z$ to $M$ via a simple spectral interpolation which will be explained in Sec.~\ref{sec:spectral:interpolation}. The interpolated $M$-band image is then passed through the encoder  $\Phi^{\text{En}}$ to obtain a new $M$-band RGB image $\bar{\vect{X}}_{\text{RGB}} \in \mathbb{R}^{H_2\times W_2\times M}$ of the desired resolution. Because the decoder is shared by two tasks, $\bar{\vect{X}}_{\text{RGB}}$ is also needed to be fed to its own task-specific decoder network $\Phi^{\text{De}}_{\text{RGB}}$ for further refinement. The final output $\hat{\vect{X}}_{\text{RGB}}$ from  $\Phi^{\text{De}}_{\text{RGB}}$ is then compared to the ground-truth image $\vect{X}_{\text{RGB}}$.   

In order to have a modular design, the three sub-networks have the same basic architecture.
They are all composed of a sequence of Spatial-Spectral Block (SSB) modules. The SBB module was proposed in~\cite{spatial:spectral:prior:20} as a basic building block for their HSI SR network. Each SBB has a Spatial Residual Module and a Spectral Attention Residual Module. Two Convolutional layers (the first one followed by a  ReLu layer) with 3x3 filters are used in the Spatial Residual Module to capture spatial correlations. Two Convolutional layers (the first one again followed by a Relu layer) with 1x1 filters are used in the Spectral Attention Residual Module to capture spectral correlations.  Please refer to Fig.2 in \cite{spatial:spectral:prior:20} for more details of the SBB module.  We construct the whole network with standard Convolutional Layers, SBBs, Upsampling Layers and Concatenation Operations. There are also skip connections at multiple scales to facilitate the information flow. The input LR images are also scaled to the desired size via Bicubic Interpolation and fused with the network output for residual learning. The complete network is shown in Fig. \ref{fig:network}. We employ the PixelShuffle \cite{pixelshuffle} operator for the upsampling layer. Given a scaling factor $\tau$, the first upsampling layer upscales the features $\tau/2$ times and the second one handles the remaining $\times2$ factor.  The internal features of all SBB modules are limited to $256$ in this work. The filter size of all Convolutional Layers, except for those in the Spectral Attention Residual Module of SBBs, are set to $3\times 3$.

\subsubsection{Spectral Interpolation of RGB Images}
\label{sec:spectral:interpolation} 
The task is to increase the number of band from $Z$ to $M$ for RGB images. For instance, we have $Z=3$ and $M=8$ in this work. Because the generated $M$-band images will be used to train the SR network for supervision transfer to HSI SR, we posit that these new images need to have certain properties. First, they should not contain artifacts. Second, the correlation between the bands of the new images should follow a distance rule in that the correlation between neighboring HSI bands should be higher than that between distant bands. For this, we propose a simple interpolation method. Given $Z$ bands, we interpolate $K=(M-Z)/(Z-1)$ new bands to each of the $Z-1$ intervals between consecutive bands. For the $i^{\text{th}}$ band $\dot{\vect{x}}(i)$ between the original bands $z$ and $z+1$, we have: 
\begin{equation}
\label{eq:interpolation}
    \dot{\vect{x}}(i) = (1-\frac{i}{K+1}) \vect{x}(z) + \frac{i}{K+1}  \vect{x}(z+1). 
\end{equation}
Note that if $K$ is not an integer, we use $\ceil*{K}$ for the first interval and $\floor*{K}$ for the second one.

\subsection{Spectral Mixup} 
\label{sec:spectralmixup}
Data augmentation is a strategy to create virtual samples by alternating the original samples. The recent \emph{mixup} method~\cite{zhang2018mixup} creates virtual examples by using convex combinations of pairs of examples and their labels. While it is very effective for high-level classification tasks, it does not offer help for low-level SR tasks \cite{data:augmentation:SR:cvpr20} because the detailed image structures can be destroyed by the mixing-up of two images. These detailed structures are important for SR tasks. Taking into account of this observation, we propose a data augmentation routine \emph{Spectral Mixup} specifically for HSI SR. It creates virtual samples and their ground truths by using convex combinations of spectral bands within self-image and within its ground-truth image, respectively. 

More specifically, given $\vect{x}_{\text{HS}}$ and it ground truth $\vect{X}_{\text{HS}}$, both with $C$ channels, we generate a mixing matrix $B \in \mathbb{R}^{C \times C}$ filled with random numbers sampled from a uniform distribution on the interval $[0, 1)$. $B$ is then row-wise normalized to make sure that the values in the projected image have the same magnitude as that of the original image. The new example and its ground truth are then created as: 
\begin{equation}
    \mathring{\vect{x}}_{\text{HS}}^{\text{(i,j)}} = \alpha\vect{x}_{\text{HS}}^{\text{(i,j)}} + (1-\alpha)B\vect{x}_{\text{HS}}^{\text{(i,j)}},
\end{equation}
\begin{equation}
     \mathring{\vect{X}}_{\text{HS}}^{\text{(i,j)}} = \alpha\vect{X}_{\text{HS}}^{\text{(i,j)}} + (1-\alpha)B\vect{X}_{\text{HS}}^{\text{(i,j)}}, 
\end{equation}
where $\text{(i,j)}$ index over all positions to get the values of pixels.
The randomly projected images are fused with the original images to strike a balance between increasing variations and preserving the fidelity of real HSIs. For instance, the relationships between the bands of real HSIs should be largely kept. In this work, $\alpha$ is set to $0.5$ and we study the influence of this parameter in Sec.~\ref{sec:exp}. The implementation of \emph{Spectral Mixup} training is very straightforward and can be done with a few lines of code. \emph{Spectral Mixup} also introduces very little computation overhead. By applying it, more examples from the vicinity of the original example can be sampled. Learning with those new examples encourages the network to have simple linear behavior in-between spectral bands which is found very useful for HSI SR.

 \begin{figure}[!tb]
    \centering
    \includegraphics[width=\linewidth]{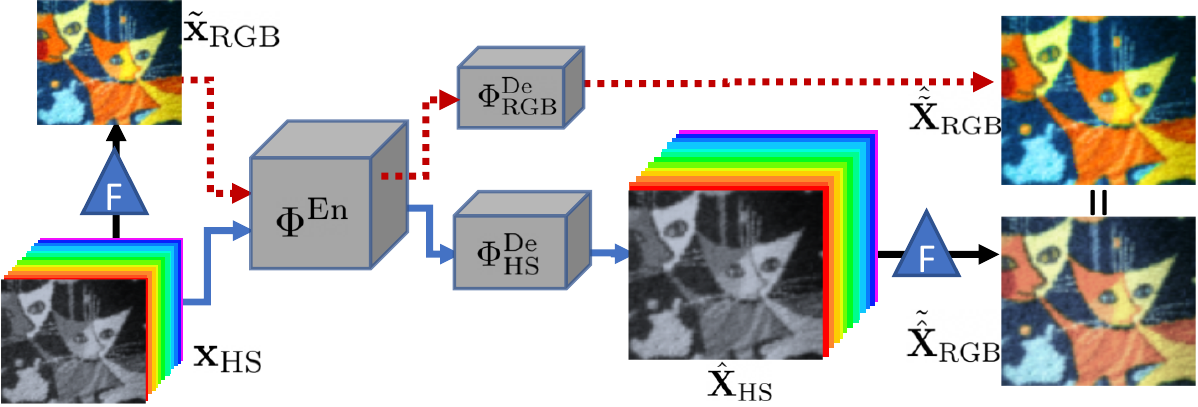}
    \caption{The pipeline of our semi-supervised learning.}
    \label{fig:ssl}
\end{figure}

\subsection{Semi-Supervised HSI SR}
\label{sec:ssl}
While training with auxiliary RGB SR task and \emph{Spectral Mixup} can greatly improve the performance, there is still a strong need to also learn with unlabeled HSIs, \ie LR HSIs without HR HSIs as ground truth. This is especially true as modern snapshot HS cameras that captures LR HSIs at high frame rate are becoming more and more accessible. In the literature, there has been a diverse sets of methods developed for semi-supervised learning (SSL) based on techniques such as entropy minimization and pseudo-labels generation. However, they are mostly designed for high-level recognition tasks and can not be applied to HSI SR directly. 

In this work, we propose a new SSL method specifically for HSI SR. For this purpose, we again leverage the fact that RGBI SR is a better-addressed problem, given that it has a large amount of training data and it predicts only three channels. The method works as follows: given an image $\vect{x}_{\text{HS}}$, we convert it to an RGB image $\tilde{\vect{x}}_{\text{RGB}}$ with the camera response function of a standard RGB camera $\vect{f}$:   $\tilde{\vect{x}}_{\text{RGB}}^{(i,j)} = \vect{f} * \vect{x}_{\text{HS}}^{(i,j)}$, where $*$ is a convolution operation. The operation is to integrate the spectra signatures into R, G, and B channels and is widely used in the literature \cite{hyperspectral:rgb:guidance:19}. The response function of Canon 1D Mark 3~\cite{cam:response} is used in this work but other camera response functions can also be used. 

The original HSI $\vect{x}_{\text{HS}}$ and the converted RGB image $\tilde{\vect{x}}_{\text{RGB}}$ are then fed into the HSI SR network $\Phi_{\text{HS}}$ and the RGBI SR network $\Phi_{\text{RGB}}$, respectively, to generate the super-resolved results:  $\hat{\vect{X}}_{\text{HS}}=\Phi_{\text{HS}}(\vect{x}_{\text{HS}})$
and $\hat{\tilde{\vect{X}}}_{\text{RGB}}=\Phi_{\text{RGB}}(\tilde{\vect{x}}_{\text{RGB}})$;  $\hat{\vect{X}}_{\text{HS}}$ is then converted to an RGB image  by using the same camera response function: $\tilde{\hat{\vect{X}}}_{\text{RGB}}^{(i,j)}=\vect{f}  * \hat{\vect{X}}_{\text{HS}}^{(i,j)}$. Finally, a consistency loss $L_{\text{ssl}}(\hat{\tilde{\vect{X}}}_{\text{RGB}},\tilde{\hat{\vect{X}}}_{\text{RGB}})$ is computed between the two HR RGB results. This consistency check makes a good use of `unlabeled' HSIs and `labeled' RGB images. It transfers supervision from the RGB side to the HSI side. The diagram of this SSL method is shown in Fig. \ref{fig:ssl}.


\subsection{Loss Function}
\label{sec:losses}
In order to capture both spatial and spectral correlation of the SR results, we follow \cite{spatial:spectral:prior:20} and combine the L1 loss and the spatial-spectral total variation (SSTV) loss \cite{sstV:tip:16}. SSTV is used to encourage smooth results in both spatial domain and spectral domain and it is defined as: 
\begin{equation}
    \mathcal{L_{\text{SSTV}}}=\frac{1}{N} \sum_{n=1}^N (||\triangledown_{\text{h}}\hat{\vect{X}}^n||_1 + ||\triangledown_{\text{w}}\hat{\vect{X}}^n||_1 + ||\triangledown_{\text{c}}\hat{\vect{X}}^n||_1),
\end{equation}
where $\triangledown_{\text{h}}$, $\triangledown_{\text{w}}$, and $\triangledown_{\text{c}}$ compute gradient along the horizontal, vertical and spectral directions, resp. The loss is: 
\begin{equation}
    \mathcal{L}= \mathcal{L}_1 +  \mathcal{L_{\text{SSTV}}}.
\end{equation}
The overall loss for our SR tasks is: 
\begin{dmath}
    \mathcal{L}^{\text{Total}}= \mathcal{L}^{\text{HS}}(\vect{X}_{\text{HS}}, \hat{\vect{X}}_{\text{HS}}) + 
      \mathcal{L}^{\text{RGB}}(\vect{X}_{\text{RGB}}, \hat{\vect{X}}_{\text{RGB}}) +
    \mathcal{L}^{\text{SMixup}}(\mathring{\vect{X}}_{\text{HS}}, \hat{\mathring{\vect{X}}}_{\text{HS}}) +
    \mathcal{L}^{\text{SSL}}(\hat{\tilde{X}}_{\text{RGB}}, \tilde{\hat{\vect{X}}}_{\text{RGB}}). 
\end{dmath}
The main loss is augmented by the three auxiliary losses which are optional but highly beneficial. A joint training with all losses together works well in principle by stacking multiple types of data samples in a single mini-batch. However, that will heavily limits the size of the training data for each loss. In this work, we adopt an alternating training strategy; that is to train with each of the four losses in turn in every iteration. In our implementation, the weights for all losses are set to $1$. The contributions of different terms are balanced or controlled by altering the number of mini-batches for that loss in each iteration. The influence of these numbers are studied in Sec.~\ref{sec:ablation}.


\section{Experiments}
\label{sec:exp}

\subsection{Experimental Setup}
\textbf{Datasets}.
We evaluate our method on four public datasets. The datasets considered are three nature HSI datasets: CAVE dataset~\cite{cave:10}, Harvard dataest~\cite{harward:dataset}, and NTIRE 2020 dataset \cite{Arad_2020_CVPR_Workshops}, and one remote sensing HSI dataset Chikusei \cite{chikusei:dataset}. Images in CAVE and NTIRE 2020 dataset have $31$ bands ranging from $400$ nm to $700$ nm at a step of $10$ nm. Images in Harvard dataset contain $31$ bands as well but range from $420$ nm to $720$ nm. 
The Chikusei dataset has $128$ bands spanning from $363$ nm to $1018$ nm.

The CAVE dataset contains $32$ images of 512 x 512 pixels. We use $20$ images for training and $10$ images for testing. We evaluated in a supervised setting and a semi-supervised setting. For our semi-supervised setting, $5$ images are used as labeled images (with HR HSIs) and the remaining $15$ used as unlabeled. For the Harvard dataset, there are $50$ images in total. We use $40$ for training and $10$ for test. For the semi-supervised setting, $6$ images are taken as the labeled images  while the remaining $34$ are taken as unlabeled images. For NTIRE 2020, there are $480$ images. We use $400$ images for training and $80$ images for test. For the semi-supervised case, we further split the $400$ images into $100$ as labeled images and $300$ as unlabeled images. For Chikusei, there is only one big image of $2517 \times 2335$ pixels. We cropped $4$ image crops of $256 \times 256$ pixels for test and use the rest for training. For the auxiliary RGBI SR task, we adopt the DIV2K Dataset \cite{AIM:dataset}. Because the resolution of DIV2K is much higher than our HSIs, we first downsample them by a factor of $\times 2$ and take these downsampled images as our HR RGB images. After cropping, it leads to $137,430$ image patches of $64 \times 64$ pixels. This is about $34$, $10$, $6$, and $40$ times larger than CAVE, Harvard  NTIRE, and  Chikusei datasets, respectively.

\begin{table*}[!tb]
  \centering
  \small
  \begin{tabular}{ccccccccccccccc}
\toprule
\#(Mini-Batches) & 0 & 1 & 2 & 3 & 4 & 5 & 6 & 8 & 10\\ \midrule
RMSE $\downarrow$ & 0.01451 & 0.01357 & 0.01329 & 0.01309 & 0.01308 & 0.01305 & 0.01315 & 0.01315 & 0.01317 \\
\bottomrule 
\end{tabular} 
\caption{Performance as a function of the number of mini-batches for RGBI SR loss.} 
  \label{tab:ablation:RGB:batchsize}  \vspace{-2mm}
\end{table*}


\begin{table}[!tb]
  \centering
  \small
  \begin{tabular}{lcccccccccccccc}
\toprule
Methods & RMSE $\downarrow$ & MPSNR $\uparrow$ & ERGAS  $\downarrow$\\ \midrule
Spatial Mixup \cite{zhang2018mixup} & 0.01308 & 41.47317 & 3.77823\\
Cutblur Mixup \cite{data:augmentation:SR:cvpr20} &0.01309 & 41.63031& 3.69937\\ 
Spectral Mixup &\textbf{0.01281} & \textbf{41.81709} & \textbf{3.64961}\\
\bottomrule 
\end{tabular} 
\caption{Ablation study for \emph{Spectral Mixup}} \vspace{-2mm}
  \label{tab:ablation:spectral mixup}
\end{table} 

\begin{table}[!tb]
  \centering
  \small
  \begin{tabular}{lcccccccccccccc}
\toprule
$M$ & 3  & 5 & 8 & 12 \\ \midrule
RMSE $\downarrow$ & 0.01321  & 0.01326 & 0.01315 & 0.01332 \\
Training time (h) & 2:14:40  & 1:49:11 & 1:24:14 & 1:15:04 \\
\bottomrule 
\end{tabular} 
\caption{Influence of group size $M$ on the performance.} 
  \label{tab:ablation:group:size} \vspace{-2mm}
\end{table} %

\textbf{Methods}. 
We compare the proposed method to four state-of-the-art HSI SR methods: GDRRN \cite{HSI:SR:grouped:recursivenet:18}, 3DFCNN \cite{3d:net:HSI:sr:17}, SSPSR \cite{spatial:spectral:prior:20}, and MCNet~\cite{2d:3d:net:HSI:SR:20}. We use the same training data for all methods and use the default training settings given by the authors of these methods. Bicubic interpolation is also introduced as a baseline. 

\textbf{Evaluation Metrics}.
We follow the literature and evaluate the performance of all methods under six standard metrics. They are cross correlation (CC) \cite{hsreview:15}, spectral angle mapper (SAM) \cite{sam:15}, root mean squared error (RMSE), erreur relative globale adimensionnelle de
synthese (ERGAS) \cite{ERGAS:02}, peak signal-to-noise ratio (PSNR),
and structure similarity (SSIM) \cite{ssim:04}. For PSNR and SSIM of
the reconstructed HSIs, their mean values of all spectral bands are reported. CC, SAM, and ERGAS are widely used in HSI fusion task, while the other three are standard metrics for image restoration and RGBI SR. Due to space limit, for some experiments, we only report numbers of three metrics and include the rest into the supplementary material.  

\textbf{Parameters}.
In this work, we focus on scaling factor $\times 4$ and $\times 8$. We report the results of $\times 4$ in the main paper, and report the results of $\times 8$ in the supplementary material. For the case of $\times 4$, we crop the images into patches of $64 \times 64$ pixels without overlapping to collect the training data. For  $\times 8$, we use patches of $128 \times 128$ pixels. Those patches are then downsampled via Bicubic interpolation to obtain the corresponding LR HSI patches.  The choice of value for other parameters are studied in Sec.~\ref{sec:ablation}.   

\textbf{Training Details}. 
We use ADAM optimizer \cite{adam} and train all variants of our method for $10$ epoches. This is a small number compared to the ones used by comparison methods. For instance, GDRRN \cite{HSI:SR:grouped:recursivenet:18} trains for $30$ epoches,  3DFCNN \cite{3d:net:HSI:sr:17} trains for $200$ epoches,  SSPSR \cite{spatial:spectral:prior:20} for $40$ epoches, and MCNet \cite{2d:3d:net:HSI:SR:20} for $200$ epoches. We choose a small number in order to thoroughly evaluate all the variants of our method. We find that $10$ epoches are sufficient to give good results for our method, and believe a larger number probably can further push the numbers up. The initial learning rate of all our methods is set to $10^{-4}$ and is reduced by a factor of $0.3$ after every $3$ epoches. As to the batch size, $16$ is used for all experiments except for the case when the SSL loss is added. For that $8$ is used due to the limit of GPU memory.

\begin{table*}[!tb]
  \centering
  \setlength\tabcolsep{1.8pt}
  \small
  \begin{tabular}{lccc|ccc|ccc|cccc}
\toprule
& \multicolumn{3} {c} {Components} & \multicolumn{3} {c} {CAVE} & \multicolumn{3} {c} {Harvard} &\multicolumn{3} {c} {NTIRE}\\
Methods & RGBSR & \emph{SMixup} & SSL & RMSE $\downarrow$ & MPSNR $\uparrow$& ERGAS $\downarrow$ & RMSE $\downarrow$ & MPSNR$\uparrow$& ERGAS $\downarrow$ & RMSE $\downarrow$ & MPSNR$\uparrow$& ERGAS $\downarrow$\\ \midrule
Ours  & & & & 0.01451 & 40.83762 & 4.03386 & 0.01406 & 40.46614 & 3.17093 &0.01602 &38.31542 & 2.20746 \\ 
Ours  & \cmark & &  & 0.01309 & 41.64471 &3.70784 &  0.01372 & 40.70069 &3.09505 & 0.01519 &38.78698 & 2.10917\\ 
Ours &  &\cmark & &  0.01353 & 41.51967 & 3.78138 &  0.01392 & 40.54614 & 3.13421 & 0.01563 & 38.58965 & 2.14927 \\
Ours  &  \cmark & \cmark& & 0.01281 & 41.81709 & 3.64961 &  0.01359 & 40.75872 & 3.07788 & 0.01526 & 38.72326 & 2.11944 \\  
Ours &  \cmark & & \cmark&  0.01254 & 42.01961 & 3.58921 &  0.01345 & 40.80595 & 3.06239 & 0.01509 & 38.83281 & 2.09676\\
Ours (final) &  \cmark &\cmark &\cmark & \textbf{0.01191} & \textbf{42.35848} & \textbf{3.44471} & \textbf{0.01331} & \textbf{40.93154} & \textbf{3.01392} & \textbf{0.01486} & \textbf{38.96572} & \textbf{2.06742}\\  \midrule
Bicubic & - & - & - & 0.01856 & 38.73800 & 5.27190 & 0.01678 &38.89758 & 3.80698 & 0.02353 &34.74012 & 3.19014\\
GDRRN~\cite{HSI:SR:grouped:recursivenet:18}  & - & - & - &  0.02463 & 36.27752 & 7.00438 &  0.01609 & 38.69532 & 4.30316 & 0.01974 & 36.07933 & 2.81752\\ 
3DFCNN~\cite{3d:net:HSI:sr:17}  & - & - & - &  0.01738 & 38.39284 & 6.70559 &  0.01578 & 39.34414 & 3.61725 & 0.02083 & 35.66309 & 2.82461 \\
SSPSR~\cite{spatial:spectral:prior:20}  & - & - & - & 0.01448 & 40.91316 & 4.04064 &  0.01427 & 40.32095 & 3.22745 & 0.01636 & 38.07401 & 2.25393\\
MCNet~\cite{2d:3d:net:HSI:SR:20}  & - & - & - &  0.01461 & 40.73858 & 4.16596 & 0.014682 & 40.18739 & 3.26059 & 0.01680 & 38.02486 & 2.28342 \\ 
\bottomrule 
\end{tabular} 
\caption{Results of all methods on the CAVE, Harvard, and NTIRE datasets in the semi-supervised setting for the $\times 4$ case. } 
  \label{tab:semi:all}
\end{table*} 

\begin{table*}[!tb]
  \centering
  \small
  \begin{tabular}{lcccccccccccccc}
\toprule
& \multicolumn{3} {c} {Components} & \multicolumn{6} {c} {Metrics} \\
Methods & RGBI SR & \emph{S. Mixup} & SSL & RMSE $\downarrow$ & CC $\uparrow$ & MPSNR $\uparrow$ & MSSIM $\uparrow$ & ERGAS $\downarrow$ & SAM $\downarrow$\\ \midrule
Ours  & & & &0.01230 & 0.94992 & 39.71319 &0.93529  & 5.32514 & 2.58381\\ 
Ours  & \cmark & & &0.01211&0.95161 &39.86037& 0.93691 &5.22631  &2.53569 \\ 
Ours &  &\cmark & & 0.01216&0.95097  &  39.82008 & 0.93649 & 5.26244 & 2.56245\\ 
Ours  &  \cmark & \cmark& &0.01215 &0.95096 & 39.8338&0.93675 &5.24407&2.59671\\ 
Ours &  \cmark & & \cmark& 0.01219&0.95109  & 39.81107 & 0.93617 & 5.25279& 2.56749\\ 
Ours (final) &  \cmark &\cmark &\cmark & \textbf{0.01181} &\textbf{0.95375}& \textbf{40.09431} & \textbf{0.94035} &\textbf{5.08513} &\textbf{2.49154} \\ 
\bottomrule 
\end{tabular} 
\caption{Results of all methods on the Chikusei dataset in the semi-supervised setting for the $\times 4$ case.} \vspace{-2mm}
  \label{tab:semi:Chikusei}
\end{table*} 

\subsection{Ablation Study}
\label{sec:ablation}
We analyze the parameter choices of our method in this section. Experiments are conducted on the CAVE dataset in the semi-supervised setting. 

\textbf{Number of mini-batches}. The number of mini-batches for labeled HSIs in each iteration is fixed to $1$. The number of mini-batches for unlabeled HSIs for the SSL is fixed to $3$. This is decided by the ratio of the size of unlabeled data to the size of labeled data. We have studied the influence of the other two parameters. First, for RGB data, we evaluate over a large range of values. The results are shown in Table~\ref{tab:ablation:RGB:batchsize}. The performance increases first with the number of mini-batches and then decreases with it. We fix the number of batches for RGB data to $3$ as it is a good trade-off between performance and computational time. We have also tested the number of mini-batches for the virtual samples created by \emph{Spectral Mixup} and find that $2$ is better than $1$ and values larger than $2$ add marginal improvement. We use $2$ for all our experiments.


\textbf{Influence of $\alpha$}. We tested five values for it: $0$, $0.25$, $0.5$, $0.75$ and $1$ and obtained the following RMSE results: $0.01175$ $0.01178$, $0.01173$, $0.01187$, and $0.01189$. Results show that $\alpha$ should not be too big or too small. We choose $\alpha=0.5$ to keep a balance between increasing data variation and preserving the fidelity of real HSIs. 

\textbf{Mixup Methods}. We compare our \emph{Spectral Mixup} to two recent data augmentation methods \emph{mixup} \cite{zhang2018mixup} and CutblurMixup \cite{data:augmentation:SR:cvpr20}. These methods also rely on the data mixing idea. Note that the results for this experiment are generated under a joint training with the auxiliary RGBI SR module as this is a stronger method than our base model. The results in Table \ref{tab:ablation:spectral mixup} show that  \emph{Spectral Mixup} outperforms both of the methods. The spatial mixing method \cite{zhang2018mixup} blends data from two images. It may break detailed structures that are important for SR tasks. The CutblurMixup method learns \emph{where to perform the SR} and is found helpful for RGB SR \cite{data:augmentation:SR:cvpr20}.  \emph{Spectral Mixup} creates virtual examples by using convex combinations of spectral bands of the same image which avoids breaking image structures and preserves simple linear behavior in-between spectral bands. This makes it more suitable for HSI tasks.

\textbf{Group Size}. We evaluated four values for $M$. The results in Table~\ref{tab:ablation:group:size} show that the performance of the method is robust to different values of $M$. It can also be found that $M=8$ gives the best result in terms of RMSE and computational time. We thus use $8$ for all our experiments. This finding is in line with \cite{spatial:spectral:prior:20}.  


\begin{figure*}[tb]
    \centering
    \subfloat{\includegraphics[width=0.19\textwidth]{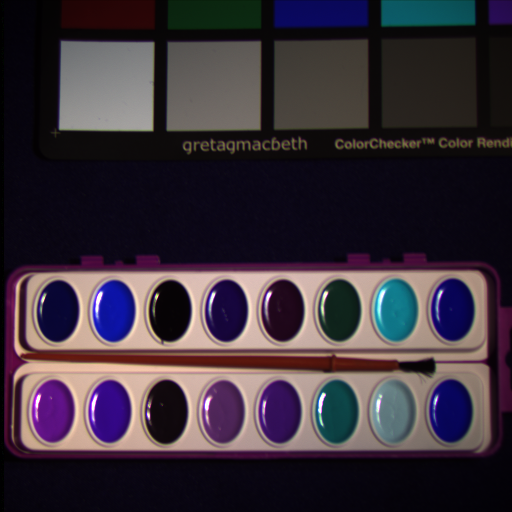}\label{fig:gt1}}
    \hfil
    \subfloat{\includegraphics[width=0.19\textwidth]{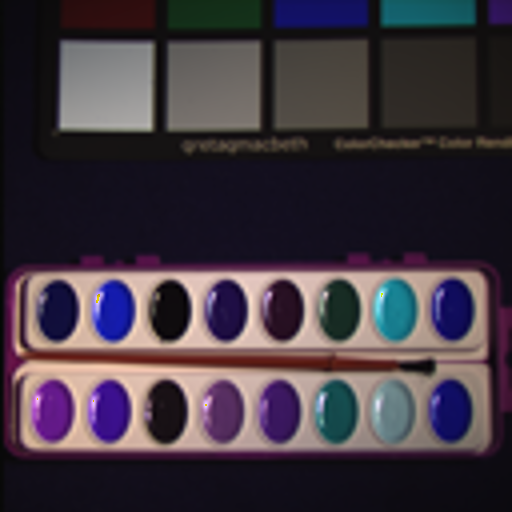}\label{fig:annotation:auxiliary}}
    \hfil
    \subfloat{\includegraphics[width=0.19\textwidth]{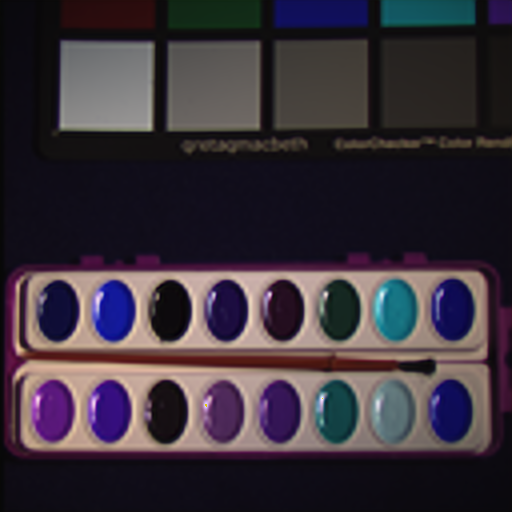}\label{fig:annotation:invalid}}
    \hfil
    \subfloat{\includegraphics[width=0.19\textwidth]{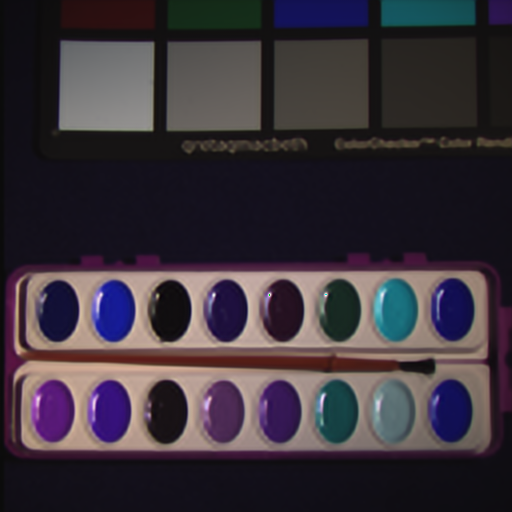}\label{fig:annotation:gt}}
        \hfil
         \subfloat{\includegraphics[width=0.19\textwidth]{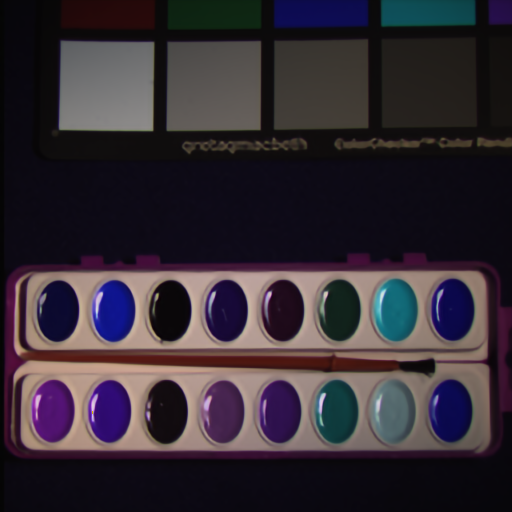}\label{fig:annotation:gt}}
        \hfil \\ \vspace{-3mm} \setcounter{subfigure}{0}
         \subfloat[Ground Truth \& Metrics]{\includegraphics[width=0.19\textwidth]{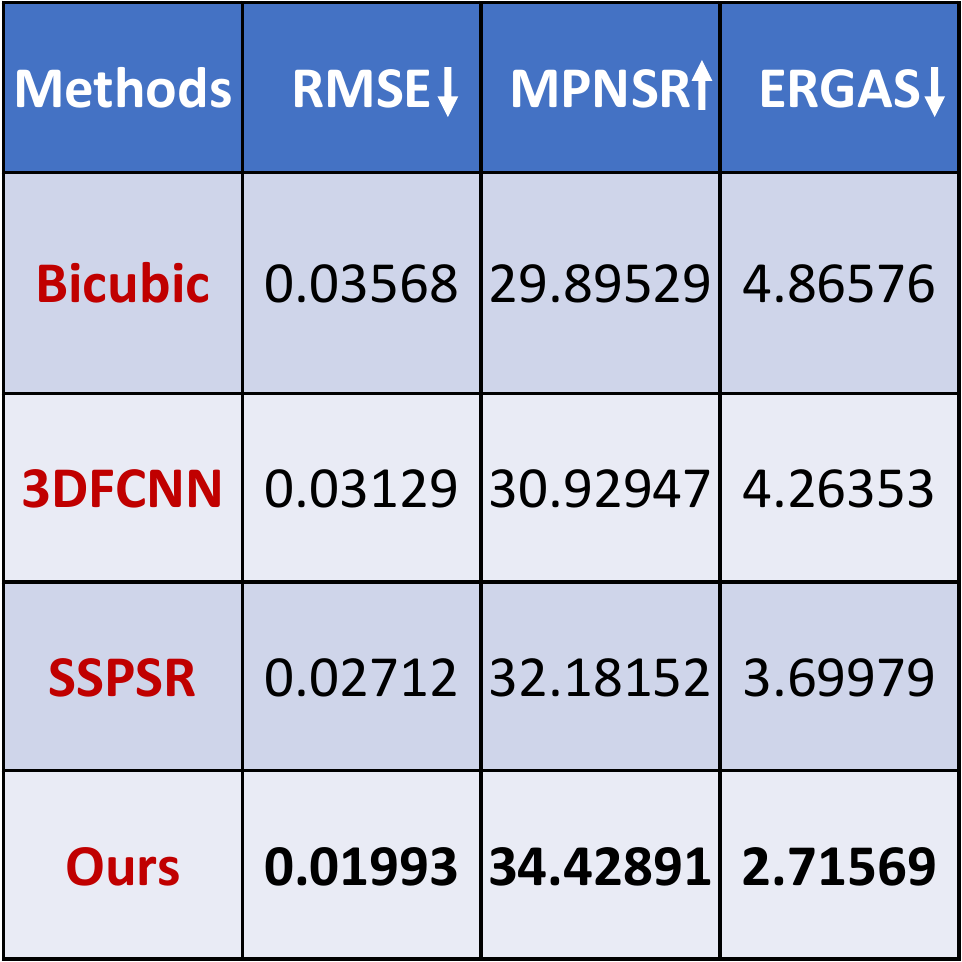}}
    \hfil
           \subfloat[Bicubic Interpolation]{\includegraphics[width=0.19\textwidth]{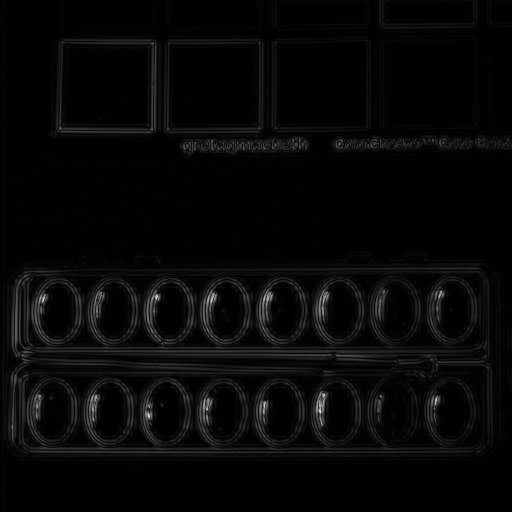}\label{fig:annotation:auxiliary}}
    \hfil
    \subfloat[3DFCNN \cite{3d:net:HSI:sr:17}]{\includegraphics[width=0.19\textwidth]{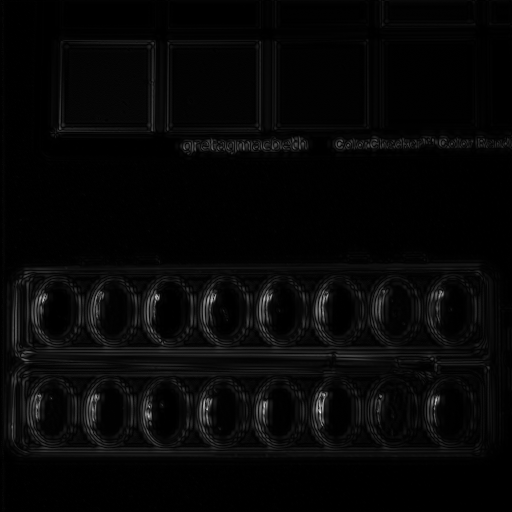}\label{fig:annotation:invalid}}
    \hfil
    \subfloat[SSPSR \cite{spatial:spectral:prior:20}]{\includegraphics[width=0.19\textwidth]{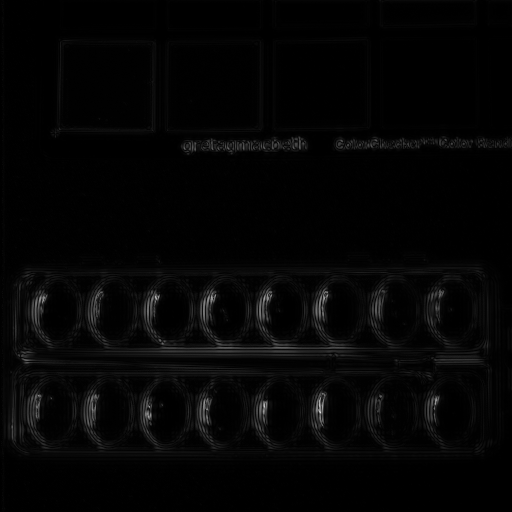}\label{fig:annotation:gt}}
        \hfil
         \subfloat[Ours]{\includegraphics[width=0.19\textwidth]{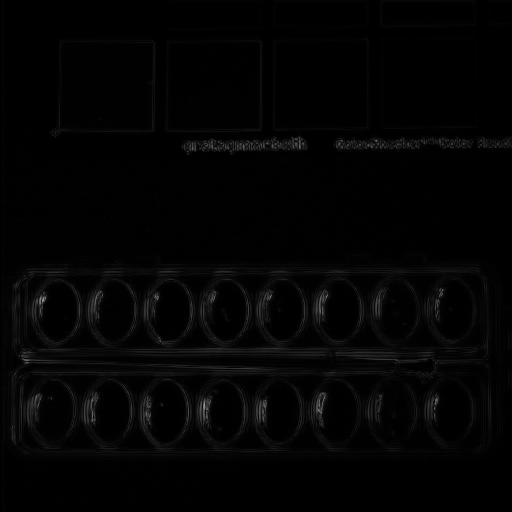}\label{fig:annotation:gt}}
        \hfil 
    \caption{Exemplar results of our method and two competing methods trained in the semi-supervised setting on the CAVE dataset: top row for the super-resolved results and bottom row for the error maps.} 
    \label{fig:result}
\end{figure*}

\begin{table*}[!tb]
  \centering
  \small
  \begin{tabular}{lcc|ccc|cccccccc}
\toprule
& \multicolumn{2} {c} {Components} & \multicolumn{3} {c} {CAVE} & \multicolumn{3} {c} {Harvard} \\
Methods & RGBI SR & \emph{S. Mixup}  & RMSE $\downarrow$ & MPSNR $\uparrow$ & ERGAS $\downarrow$ & RMSE $\downarrow$ & MPSNR $\uparrow$ & ERGAS $\downarrow$ \\ \midrule
Ours  & & &  0.01196 & 42.38359 & 3.45903 & 0.01344 & 40.91014 & 3.01039  \\ 
Ours  & \cmark &  & 0.01109 & 42.73668 & 3.35884 & 0.01325 & 41.03709 & 2.96643\\ 
Ours &  &\cmark &  0.01134 & 42.88402 & 3.28051 & \textbf{0.01317} & \textbf{41.08568} &  \textbf{2.95718} \\
Ours  &  \cmark & \cmark&  \textbf{0.01068} & \textbf{43.32421} & \textbf{3.11799} & 0.01321 & 41.05925 &  2.96496 \\ \midrule 
GDRRN~\cite{HSI:SR:grouped:recursivenet:18}  &  - & - &  0.01629 & 39.74705 & 4.52683 & 0.01484 & 39.62759 & 3.67932 \\ 
3DFCNN~\cite{3d:net:HSI:sr:17}  & - & -  & 0.01583 &39.21786  & 5.41798 & 0.01519 & 39.66271 & 3.47738  \\
SSPSR~\cite{spatial:spectral:prior:20}  & - & - & 0.01245 & 42.13787 & 3.55146 & 0.01352 & 40.81499 &3.05007  \\
MCNet~\cite{2d:3d:net:HSI:SR:20}  & - & - &  0.01245 & 42.25978 & 3.56246 & 0.01405 & 40.59229 &  3.10529 \\ 
\bottomrule 
\end{tabular} 
\caption{Results of all methods on the CAVE and Harvard datasets in the fully-supervised setting for the $\times 4$ case. } 
  \label{tab:full}
\end{table*} %

\subsection{Main Results}
\label{sec:main:results}

We first present the results in the semi-supervised setting. The results of all competing methods and all variants of our method on the CAVE, Harvard, and NTRIE  dataset are shown in Table \ref{tab:semi:all}. The results in this table and other results in supplemental material show that our method outperforms all other state-of-the-art methods significantly and consistently over all datasets and under all evaluation metrics.  We would like to point out that our baseline model -- our method without any of the three proposed contributions -- is already a top-performing method and performs better than other comparing methods. 

The good performance of our base model is mainly due to the use of deep group convolutional networks for this task.  Our results reinforce the findings made in \cite{spatial:spectral:prior:20} that group convolutional networks are good at extracting the correlation between spectral bands without increasing the model size. The network of~\cite{3d:net:HSI:sr:17} is quite shallow, probably because 3D convolution based methods are computationally heavy in general. We believe that this is the reason why their method does not give top results. When compared to the very recent method MCNet \cite{2d:3d:net:HSI:SR:20}, our base model also performs better in almost all cases. This is especially interesting because MCNet is trained for $200$ epoches while our method is trained only for $10$ epoches.  

 
The proposed contributions, namely training with the auxiliary task RGBI SR, data augmentation via \emph{Spectral Mixup} and the semi-supervised learning method based on cross-model consistency, all contribute significantly to the final results. Out of the three, learning with the auxiliary task RGBI SR and \emph{Spectral Mixup} can work on its own. The SSL component needs to be used together with the auxiliary task RGBI SR. The results show that our SSL method can provide further improvement on top of the auxiliary RGBI SR method, and when the three components are combined together, we get the best performance -- better than using any subsets of the proposed contributions. These observations are also well supported by our results on Chikusei with 128 bands in Table \ref{tab:semi:Chikusei}.

When more supervision is given such as in the fully-supervised setting, all conclusions we drawn in the semi-supervised setting hold as shown in Table \ref{tab:full}. The results show that the proposed components are very effective and can be applied to situations with varying amount of HR HSIs. We show visual results of our  method and two competing methods in Fig.~\ref{fig:result}. The figure shows that our method generate few errors. More results can be found in the supplementary material. 

\subsection{Discussion}
The superior performance shows that our method is able to learn from heterogeneous datasets and virtual examples rather than from purely HR HSIs. This greatly increases the amount of training data that can be used for HSI SR and can also include training samples for scenes such as moving objects that cannot be captured easily with the current hyperspectal imaging devices. We would like to point out that while the general concepts of learning with auxiliary tasks, data augmentation, and semi-supervised learning are well known, the challenge and novelty lie in defining proper auxiliary tasks for a new main task, proposing suitable augmentation technique and developing effective SSL methods for a new task. For instance, many seemingly-related auxiliary tasks yield no improvement or even degrade the performance of the main task \cite{auxiliary:task:nips20}. 
-

\section{Conclusion} 
\label{sec:conclusion}

In this paper, we have proposed a new method for hypserspectral image (HSI) super-resolution (SR). We build a deep group convolutional network which yields the state-of-the-art results. To further improve it, we have proposed three contributions. First, we extend the network such that the HSI SR task can be trained together with an auxiliary RGB image SR task to gain more supervision. Second, a simple, yet effective data augmentation method \emph{Spectral Mixup} is proposed to create virtual training samples for HSI SR to increase the robustness of the network to new examples. Finally, the network is extended to also learn from datasets with LR HSIs only. The contributions greatly increase the amount of training data that HSI SR methods can use. Extensive experiments show that all the three contributions are important and they help our method set a new state of the art on four public datasets. 

{\small
\bibliographystyle{ieee_fullname}
\bibliography{egbib}

\begin{thebibliography}{10}\itemsep=-1pt

\bibitem{sstV:tip:16}
H.~K. {Aggarwal} and A. {Majumdar}.
\newblock Hyperspectral image denoising using spatio-spectral total variation.
\newblock {\em IEEE Geoscience and Remote Sensing Letters}, 13(3):442--446,
  2016.

\bibitem{AIM:dataset}
E. {Agustsson} and R. {Timofte}.
\newblock Ntire 2017 challenge on single image super-resolution: Dataset and
  study.
\newblock In {\em IEEE Conference on Computer Vision and Pattern Recognition
  Workshops (CVPRW)}, 2017.

\bibitem{HSI:SR:05}
T. {Akgun}, Y. {Altunbasak}, and R.~M. {Mersereau}.
\newblock Super-resolution reconstruction of hyperspectral images.
\newblock {\em IEEE Transactions on Image Processing}, 14(11):1860--1875, 2005.

\bibitem{HSISR:sparse:eccv14}
Naveed Akhtar, Faisal Shafait, and Ajmal Mian.
\newblock Sparse spatio-spectral representation for hyperspectral image
  super-resolution.
\newblock In {\em ECCV}, 2014.

\bibitem{bayesian:HSISR:15}
N. {Akhtar}, F. {Shafait}, and A. {Mian}.
\newblock Bayesian sparse representation for hyperspectral image super
  resolution.
\newblock In {\em CVPR}, 2015.

\bibitem{HSISR:beta:eccv16}
Naveed Akhtar, Faisal Shafait, and Ajmal Mian.
\newblock Hierarchical beta process with gaussian process prior for
  hyperspectral image super resolution.
\newblock In Bastian Leibe, Jiri Matas, Nicu Sebe, and Max Welling, editors,
  {\em ECCV}, 2016.

\bibitem{arad_and_ben_shahar_2016_ECCV}
Boaz Arad and Ohad Ben-Shahar.
\newblock Sparse recovery of hyperspectral signal from natural rgb images.
\newblock In {\em ECCV}, 2016.

\bibitem{Arad_2020_CVPR_Workshops}
Boaz Arad, Radu Timofte, Ohad Ben-Shahar, Yi-Tun Lin, and Graham~D. Finlayson.
\newblock Ntire 2020 challenge on spectral reconstruction from an rgb image.
\newblock In {\em Proceedings of the IEEE/CVF Conference on Computer Vision and
  Pattern Recognition (CVPR) Workshops}, June 2020.

\bibitem{semanticbinaural}
Arun {Balajee Vasudevan}, Dengxin Dai, and Luc {Van Gool}.
\newblock Semantic object prediction and spatial sound prediction with binaural
  sounds.
\newblock In {\em European Conference on Computer Vision (ECCV)}, 2020.

\bibitem{s4l:iccv19}
Lucas Beyer, Xiaohua Zhai, Avital Oliver, and Alexander Kolesnikov.
\newblock {S4L:} self-supervised semi-supervised learning.
\newblock In {\em ICCV}, 2019.

\bibitem{unmixing:survey:12}
J.~M. {Bioucas-Dias}, A. {Plaza}, N. {Dobigeon}, M. {Parente}, Q. {Du}, P.
  {Gader}, and J. {Chanussot}.
\newblock Hyperspectral unmixing overview: Geometrical, statistical, and sparse
  regression-based approaches.
\newblock {\em IEEE Journal of Selected Topics in Applied Earth Observations
  and Remote Sensing}, 5(2):354--379, 2012.

\bibitem{harward:dataset}
A. Chakrabarti and T. Zickler.
\newblock {Statistics of Real-World Hyperspectral Images}.
\newblock In {\em CVPR}, 2011.

\bibitem{data:fusion:15}
C. {Chen}, Y. {Li}, W. {Liu}, and J. {Huang}.
\newblock Sirf: Simultaneous satellite image registration and fusion in a
  unified framework.
\newblock {\em IEEE Transactions on Image Processing}, 24(11):4213--4224, 2015.

\bibitem{mtl:loss:cvpr18}
R. {Cipolla}, Y. {Gal}, and A. {Kendall}.
\newblock Multi-task learning using uncertainty to weigh losses for scene
  geometry and semantics.
\newblock In {\em CVPR}, 2018.

\bibitem{Dai_2013_ICCV}
Dengxin Dai and Luc Van~Gool.
\newblock Ensemble projection for semi-supervised image classification.
\newblock In {\em ICCV}, 2013.

\bibitem{HSI:SR:tensor:17}
R. {Dian}, L. {Fang}, and S. {Li}.
\newblock Hyperspectral image super-resolution via non-local sparse tensor
  factorization.
\newblock In {\em CVPR}, pages 3862--3871, 2017.

\bibitem{sr:eccv14}
C. {Dong}, C.~C. {Loy}, K. {He}, and X. {Tang}.
\newblock Image super-resolution using deep convolutional networks.
\newblock {\em IEEE Transactions on Pattern Analysis and Machine Intelligence},
  38(2):295--307, 2016.

\bibitem{HSI:sparse:coding:tip16}
W. {Dong}, F. {Fu}, G. {Shi}, X. {Cao}, J. {Wu}, G. {Li}, and X. {Li}.
\newblock Hyperspectral image super-resolution via non-negative structured
  sparse representation.
\newblock {\em IEEE Transactions on Image Processing}, 25(5):2337--2352, 2016.

\bibitem{hyperspectral:rgb:guidance:19}
Y. {Fu}, T. {Zhang}, Y. {Zheng}, D. {Zhang}, and H. {Huang}.
\newblock Hyperspectral image super-resolution with optimized rgb guidance.
\newblock In {\em CVPR}, 2019.

\bibitem{Galliani2017LearnedSS}
S. Galliani, Charis Lanaras, D. Marmanis, E. Baltsavias, and K. Schindler.
\newblock Learned spectral super-resolution.
\newblock {\em ArXiv}, abs/1703.09470, 2017.

\bibitem{remote:sensing:09}
Alexander~F.H. Goetz.
\newblock Three decades of hyperspectral remote sensing of the earth: A
  personal view.
\newblock {\em Remote Sensing of Environment}, 113:S5 -- S16, 2009.

\bibitem{food:review:07}
A.A. Gowen, C.P. O'Donnell, P.J. Cullen, G. Downey, and J.M. Frias.
\newblock Hyperspectral imaging – an emerging process analytical tool for
  food quality and safety control.
\newblock {\em Trends in Food Science \& Technology}, 18(12):590 -- 598, 2007.

\bibitem{guo2020closed}
Yong Guo, Jian Chen, Jingdong Wang, Qi Chen, Jiezhang Cao, Zeshuai Deng, Yanwu
  Xu, and Mingkui Tan.
\newblock Closed-loop matters: Dual regression networks for single image
  super-resolution.
\newblock In {\em CVPR}, 2020.

\bibitem{hoyer2020three}
Lukas Hoyer, Dengxin Dai, Yuhua Chen, Adrian Köring, Suman Saha, and Luc
  Van~Gool.
\newblock Three ways to improve semantic segmentation with self-supervised
  depth estimation.
\newblock {\em arXiv preprint arXiv:2012.10782}, 2020.

\bibitem{cam:response}
J. {Jiang}, D. {Liu}, J. {Gu}, and S. {Süsstrunk}.
\newblock What is the space of spectral sensitivity functions for digital color
  cameras?
\newblock In {\em 2013 IEEE Workshop on Applications of Computer Vision
  (WACV)}, 2013.

\bibitem{spatial:spectral:prior:20}
J. {Jiang}, H. {Sun}, X. {Liu}, and J. {Ma}.
\newblock Learning spatial-spectral prior for super-resolution of hyperspectral
  imagery.
\newblock {\em IEEE Transactions on Computational Imaging}, 6:1082--1096, 2020.

\bibitem{very:deep:SR:16}
J. {Kim}, J.~K. {Lee}, and K.~M. {Lee}.
\newblock Accurate image super-resolution using very deep convolutional
  networks.
\newblock In {\em CVPR}, 2016.

\bibitem{adam}
Diederik~P. Kingma and Jimmy Ba.
\newblock Adam: {A} method for stochastic optimization.
\newblock In {\em ICLR}, 2015.

\bibitem{ssl:gan:14}
Durk~P Kingma, Shakir Mohamed, Danilo Jimenez~Rezende, and Max Welling.
\newblock Semi-supervised learning with deep generative models.
\newblock In {\em NeurIPS}. 2014.

\bibitem{deep:laplacian:cvpr17}
W. {Lai}, J. {Huang}, N. {Ahuja}, and M. {Yang}.
\newblock Deep laplacian pyramid networks for fast and accurate
  super-resolution.
\newblock In {\em CVPR}, 2017.

\bibitem{temporal:ensembling}
Samuli Laine and Timo Aila.
\newblock Temporal ensembling for semi-supervised learning.
\newblock In {\em ICLR}, 2017.

\bibitem{HSI:unmixing:iccv15}
C. {Lanaras}, E. {Baltsavias}, and K. {Schindler}.
\newblock Hyperspectral super-resolution by coupled spectral unmixing.
\newblock In {\em 2015 IEEE International Conference on Computer Vision
  (ICCV)}, 2015.

\bibitem{Ledig_2017_CVPR}
Christian Ledig, Lucas Theis, Ferenc Huszar, Jose Caballero, Andrew Cunningham,
  Alejandro Acosta, Andrew Aitken, Alykhan Tejani, Johannes Totz, Zehan Wang,
  and Wenzhe Shi.
\newblock Photo-realistic single image super-resolution using a generative
  adversarial network.
\newblock In {\em CVPR}, 2017.

\bibitem{2d:3d:net:HSI:SR:20}
Qiang Li, Qi Wang, and Xuelong Li.
\newblock Mixed 2d/3d convolutional network for hyperspectral image
  super-resolution.
\newblock {\em Remote Sensing}, 12(10), 2020.

\bibitem{HSI:SR:grouped:recursivenet:18}
Y. Li, Lei Zhang, C. Ding, Wei Wei, and Y. Zhang.
\newblock Single hyperspectral image super-resolution with grouped deep
  recursive residual network.
\newblock {\em 2018 IEEE Fourth International Conference on Multimedia Big Data
  (BigMM)}, 2018.

\bibitem{hsreview:15}
L. {Loncan}, L.~B. {de Almeida}, J.~M. {Bioucas-Dias}, X. {Briottet}, J.
  {Chanussot}, N. {Dobigeon}, S. {Fabre}, W. {Liao}, G.~A. {Licciardi}, M.
  {Simões}, J. {Tourneret}, M.~A. {Veganzones}, G. {Vivone}, Q. {Wei}, and N.
  {Yokoya}.
\newblock Hyperspectral pansharpening: A review.
\newblock {\em IEEE Geoscience and Remote Sensing Magazine}, 3(3):27--46, 2015.

\bibitem{medical:review:14}
Guolan Lua and Baowei Fei.
\newblock Medical hyperspectral imaging: a review.
\newblock {\em Journal of Biomedical Optics}, 2014.

\bibitem{3d:net:HSI:sr:17}
Shaohui Mei, Xin Yuan, Jingyu Ji, Yifan Zhang, Shuai Wan, and Qian Du.
\newblock Hyperspectral image spatial super-resolution via 3d full
  convolutional neural network.
\newblock {\em Remote Sensing}, 9(11), 2017.

\bibitem{hyspectraobjdetection}
N.~M. {Nasrabadi}.
\newblock Hyperspectral target detection : An overview of current and future
  challenges.
\newblock {\em IEEE Signal Processing Magazine}, 31(1):34--44, 2014.

\bibitem{Training:Spectral:rgb:14}
Rang M.~H. Nguyen, Dilip~K. Prasad, and Michael~S. Brown.
\newblock Training-based spectral reconstruction from a single rgb image.
\newblock In {\em ECCV}, 2014.

\bibitem{unsueprvised:fusion:cvpr18}
Ying Qu, Hairong Qi, and Chiman Kwan.
\newblock Unsupervised sparse dirichlet-net for hyperspectral image
  super-resolution.
\newblock In {\em CVPR}, 2018.

\bibitem{auxiliary:task:nips20}
Baifeng Shi, Judy Hoffman, Kate Saenko, Trevor Darrell, and Huijuan Xu.
\newblock Auxiliary task reweighting for minimum-data learning.
\newblock In {\em NeurIPS}, 2020.

\bibitem{pixelshuffle}
W. {Shi}, J. {Caballero}, F. {Huszár}, J. {Totz}, A.~P. {Aitken}, R. {Bishop},
  D. {Rueckert}, and Z. {Wang}.
\newblock Real-time single image and video super-resolution using an efficient
  sub-pixel convolutional neural network.
\newblock In {\em CVPR}, 2016.

\bibitem{HSCNN+}
Z. {Shi}, C. {Chen}, Z. {Xiong}, D. {Liu}, and F. {Wu}.
\newblock Hscnn+: Advanced cnn-based hyperspectral recovery from rgb images.
\newblock In {\em IEEE/CVF Conference on Computer Vision and Pattern
  Recognition Workshops (CVPRW)}, 2018.

\bibitem{mean:teacher:nips17}
Antti Tarvainen and Harri Valpola.
\newblock Mean teachers are better role models: Weight-averaged consistency
  targets improve semi-supervised deep learning results.
\newblock 2017.

\bibitem{mtl:survey:20}
Simon Vandenhende, Stamatios Georgoulis, Marc Proesmans, Dengxin Dai, , and Luc
  {Van Gool}.
\newblock Multi-task learning for dense prediction tasks: A survey, 2020.

\bibitem{ERGAS:02}
L. Wald.
\newblock Data fusion: definitions and architectures: fusion of images of
  different spatial resolutions.
\newblock In {\em Presses des MINES}, 2002.

\bibitem{noisy:student:20}
Qizhe Xie, Minh-Thang Luong, Eduard Hovy, and Quoc~V. Le.
\newblock Self-training with noisy student improves imagenet classification.
\newblock In {\em CVPR}, 2020.

\bibitem{fusion:net:cvpr19}
Qi Xie, Minghao Zhou, Qian Zhao, Deyu Meng, Wangmeng Zuo, and Zongben Xu.
\newblock Multispectral and hyperspectral image fusion by {MS/HS} fusion net.
\newblock In {\em CVPR}, 2019.

\bibitem{cave:10}
F. {Yasuma}, T. {Mitsunaga}, D. {Iso}, and S.~K. {Nayar}.
\newblock Generalized assorted pixel camera: Postcapture control of resolution,
  dynamic range, and spectrum.
\newblock {\em IEEE Transactions on Image Processing}, 19(9):2241--2253, 2010.

\bibitem{HSISR:fusion:survey:17}
N. {Yokoya}, C. {Grohnfeldt}, and J. {Chanussot}.
\newblock Hyperspectral and multispectral data fusion: A comparative review of
  the recent literature.
\newblock {\em IEEE Geoscience and Remote Sensing Magazine}, 5(2):29--56, 2017.

\bibitem{chikusei:dataset}
N. Yokoya and A. Iwasaki.
\newblock Airborne hyperspectral data over chikusei.
\newblock {\em Space Application Laboratory, University of Tokyo, Japan, Tech.
  Rep. SAL-2016-05-27}, May 2016.

\bibitem{data:augmentation:SR:cvpr20}
Jaejun Yoo, Namhyuk Ahn, and Kyung-Ah Sohn.
\newblock Rethinking data augmentation for image super-resolution: A
  comprehensive analysis and a new strategy.
\newblock In {\em CVPR}, 2020.

\bibitem{HSI:SR:TL:17}
Y. {Yuan}, X. {Zheng}, and X. {Lu}.
\newblock Hyperspectral image superresolution by transfer learning.
\newblock {\em IEEE Journal of Selected Topics in Applied Earth Observations
  and Remote Sensing}, 10(5):1963--1974, 2017.

\bibitem{sam:15}
R.~H. Yuhas, A.~F. Goetz, and booktitle={JPL, Summaries of the Third Annual JPL
  Airborne Geoscience Workshop} year =~{1992} J.~W.~Boardman,
  title={Discrimination among semi-arid landscape endmembers using the spectral
  angle mapper (sam) algorithm}.

\bibitem{zamir2020consistency}
Amir Zamir, Alexander Sax, Teresa Yeo, Oğuzhan Kar, Nikhil Cheerla, Rohan
  Suri, Zhangjie Cao, Jitendra Malik, and Leonidas Guibas.
\newblock Robust learning through cross-task consistency.
\newblock In {\em CVPR}. 2020.

\bibitem{zhang2018mixup}
Hongyi Zhang, Moustapha Cisse, Yann~N. Dauphin, and David Lopez-Paz.
\newblock mixup: Beyond empirical risk minimization.
\newblock In {\em International Conference on Learning Representations}, 2018.

\bibitem{Zhang_2020_CVPR}
Lei Zhang, Jiangtao Nie, Wei Wei, Yanning Zhang, Shengcai Liao, and Ling Shao.
\newblock Unsupervised adaptation learning for hyperspectral imagery
  super-resolution.
\newblock In {\em CVPR}, 2020.

\bibitem{residual:dense:sr:cvpr18}
Yulun Zhang, Yapeng Tian, Yu Kong, Bineng Zhong, and Yun Fu.
\newblock Residual dense network for image super-resolution.

\bibitem{HIS:SR:2011}
Y. Zhao, Jinxiang Yang, Qingyong Zhang, L. Song, Y. Cheng, and Q. Pan.
\newblock Hyperspectral imagery super-resolution by sparse representation and
  spectral regularization.
\newblock {\em EURASIP Journal on Advances in Signal Processing}, 2011:1--10,
  2011.

\bibitem{fusion:20}
Y. {Zhou}, A. {Rangarajan}, and P.~D. {Gader}.
\newblock An integrated approach to registration and fusion of hyperspectral
  and multispectral images.
\newblock {\em IEEE Transactions on Geoscience and Remote Sensing},
  58(5):3020--3033, 2020.

\bibitem{ssim:04}
{Zhou Wang}, A.~C. {Bovik}, H.~R. {Sheikh}, and E.~P. {Simoncelli}.
\newblock Image quality assessment: from error visibility to structural
  similarity.
\newblock {\em IEEE Transactions on Image Processing}, 13(4):600--612, 2004.

\end{thebibliography}
}

\section{Supplementary Material}
\subsection{Further results for scaling factor $\times 4$}
Due to space limitation, only results under three metrics, \ie, RMSE, MPSNR, and ERGAS, are reported in the main paper. Here, we report the results under all six considered metrics, \ie, 
RMSE, CC, MPSNR, MSSIM, ERGAS, and SAM. For the case of scaling factor $\times 4$ and in the semi-supervised setting, the results on the CAVE dataset, the Harvard dataset, the NTIRE2020 dataset are shown in Table \ref{tab:semi:cave}, Table \ref{tab:semi:harvard} and Table \ref{tab:semi:NTIRE2020}, respectively. The results under the full-supervision setting on the CAVE dataset and on the Harvard dataset are reported in Table \ref{tab:fullsup:cave} and Table \ref{tab:fullsup:harvard} \footnote{The results on the NTIRE2020 dataset have not completed before the deadline unfortunately, but the obtained results show that the variants of our method with only a subset of our contributions already outperform all the competing methods.}. 

These tables show that our method outperforms other comparison methods by a large margin under all six metrics. All our three contributions are useful and their combination yields the best results. The conclusions we have in the main paper hold for all the six metrics.

\subsection{Results for scaling factor $\times 8$}
We also provide results for scaling factor $\times 8$. The results on the CAVE, the Harvard, and the NTIRE2020 datasets in the semi-supervised setting are shown in Table \ref{tab:semi:cave:x8}, Table \ref{tab:semi:harvard:x8}, and Table \ref{tab:semi:NTIRE2020:x8}, respectively. It is evident from these tables that our method also outperforms other methods significantly and consistently for scaling factor $\times 8$. The same trend is observed for both $\times x4$ and $\times x8$ that all our three contributions are useful and their combination yields the best results. This demonstrates the applicability of our method across different scaling factors.

\subsection{More visual results}
\begin{figure*}[tb]
    \centering
    \subfloat[Ground Truth]{\includegraphics[width=0.23\textwidth]{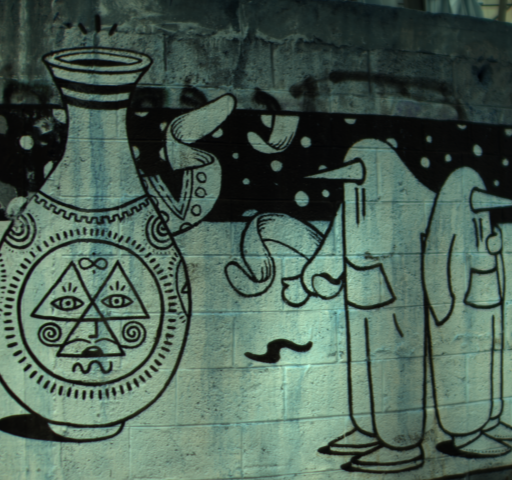}\label{fig:gt1}}
    \hfil \\
    \subfloat[Bicubic Interpolation ]{\includegraphics[width=0.23\textwidth]{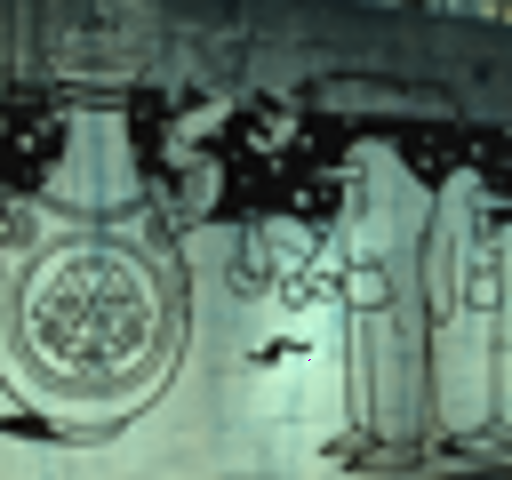}\label{fig:annotation:invalid}}
    \hfil
    \subfloat[Error of Bicubic]{\includegraphics[width=0.23\textwidth]{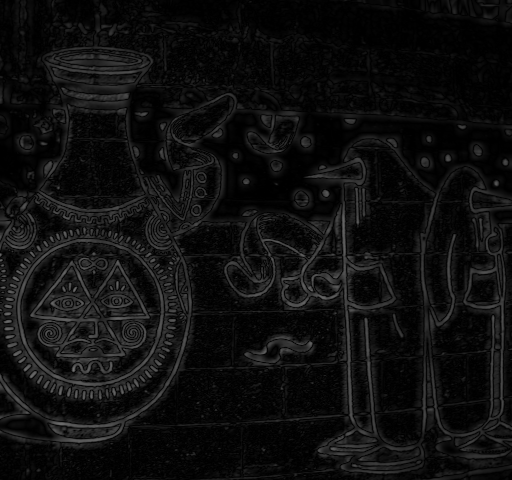}\label{fig:annotation:gt}}
        \hfil 
         \subfloat[GDRRN \cite{HSI:SR:grouped:recursivenet:18}]{\includegraphics[width=0.23\textwidth]{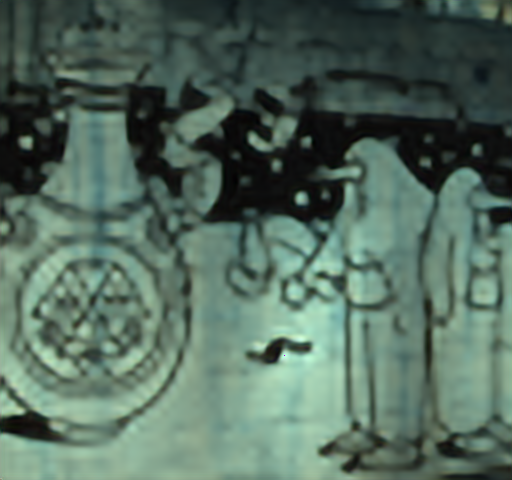}\label{fig:annotation:gt}}
        \hfil 
         \subfloat[Error of GDRRN]{\includegraphics[width=0.23\textwidth]{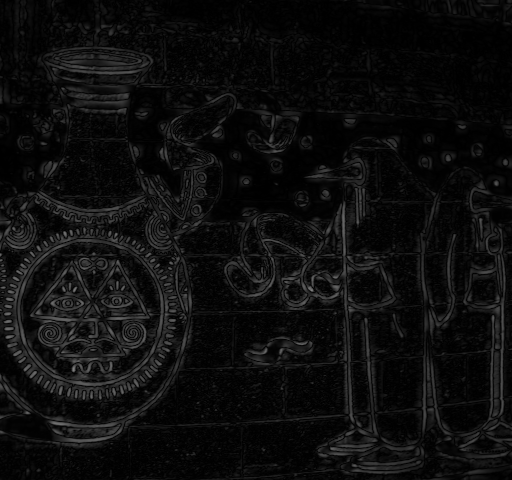}}
    \hfil  \\
           \subfloat[3DFCNN \cite{3d:net:HSI:sr:17}]{\includegraphics[width=0.23\textwidth]{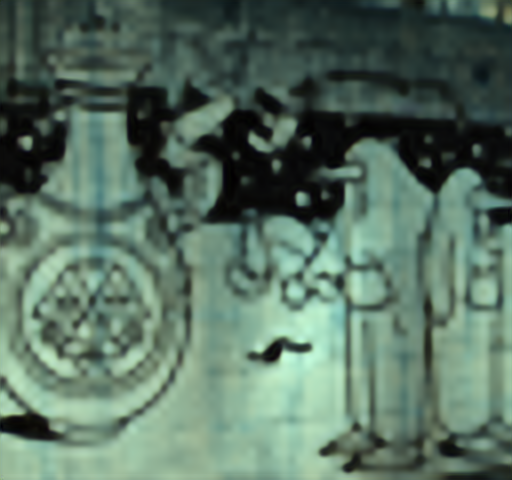}\label{fig:annotation:auxiliary}}
    \hfil
    \subfloat[Error of 3DFCNN]{\includegraphics[width=0.23\textwidth]{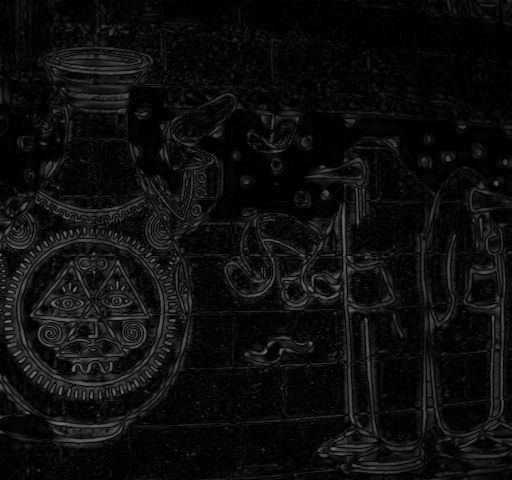}\label{fig:annotation:invalid}}
    \hfil 
    \subfloat[SSPSR \cite{spatial:spectral:prior:20}]{\includegraphics[width=0.23\textwidth]{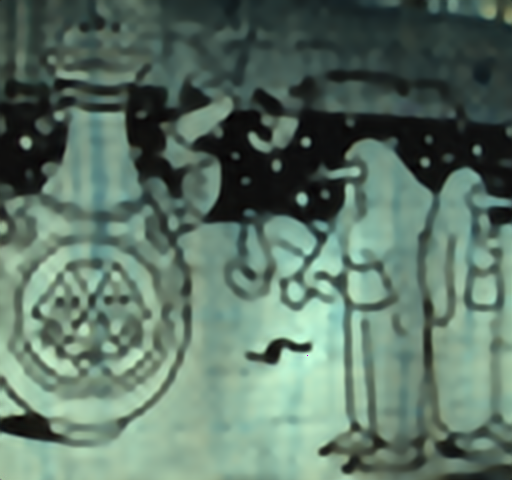}\label{fig:annotation:gt}}
        \hfil
         \subfloat[Error of SSSPSR]{\includegraphics[width=0.23\textwidth]{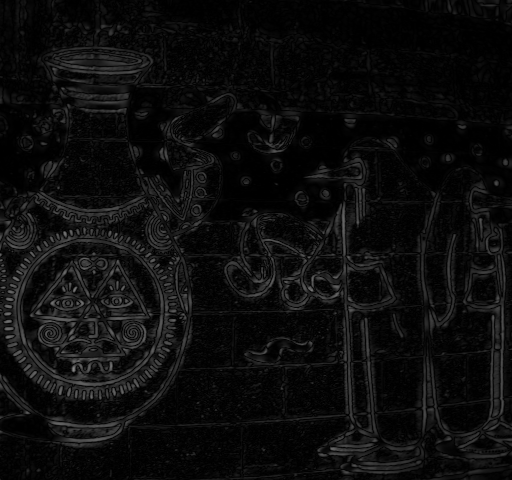}\label{fig:annotation:gt}}
        \hfil \\
    \subfloat[MCNet \cite{2d:3d:net:HSI:SR:20}]{\includegraphics[width=0.23\textwidth]{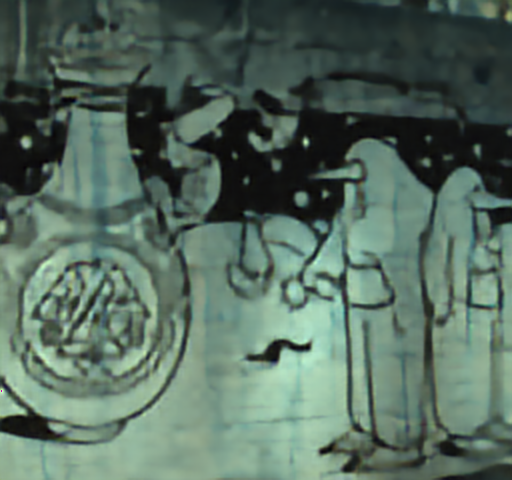}\label{fig:annotation:gt}}
        \hfil
         \subfloat[Error of MCNet]{\includegraphics[width=0.23\textwidth]{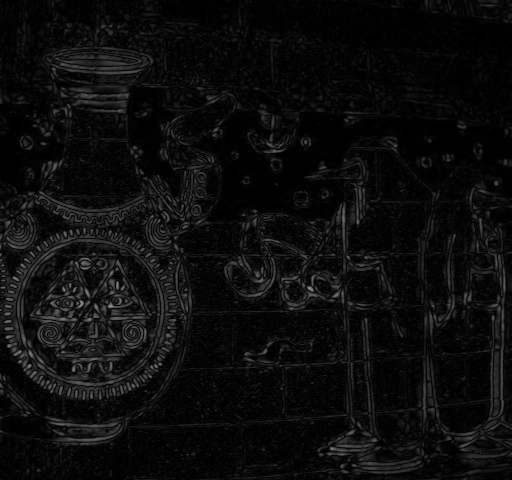}\label{fig:annotation:gt}}
        \hfil 
    \subfloat[Ours]{\includegraphics[width=0.23\textwidth]{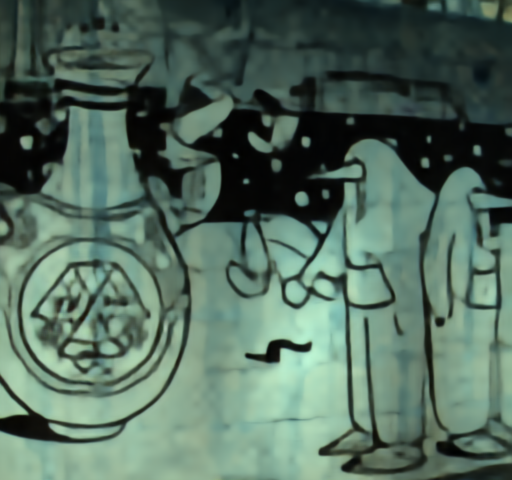}\label{fig:annotation:gt}}
        \hfil
         \subfloat[Error of Ours]{\includegraphics[width=0.23\textwidth]{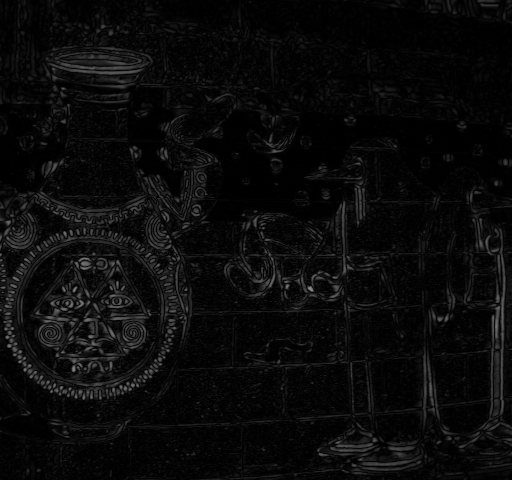}\label{fig:annotation:gt}}
        \hfil 
    \caption{Exemplar results of $\times 8$ by our method and all comparison methods. The error is L2 distance to the ground-truth pixel values, averaged over the three bands.}
    \label{fig:result1}
\end{figure*}

\begin{figure*}[tb]
    \centering
    \subfloat[Ground Truth]{\includegraphics[width=0.23\textwidth]{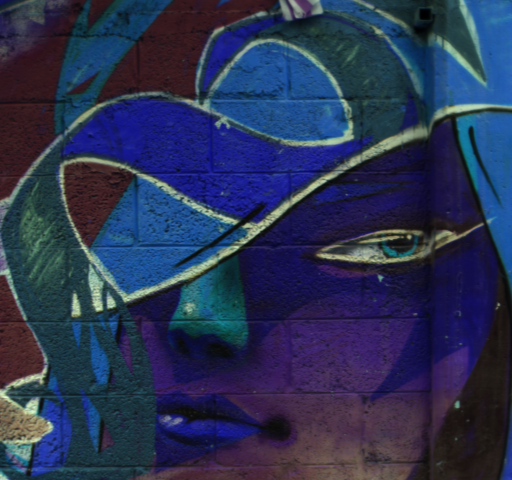}\label{fig:gt1}}
    \hfil \\
    \subfloat[Bicubic Interpolation ]{\includegraphics[width=0.23\textwidth]{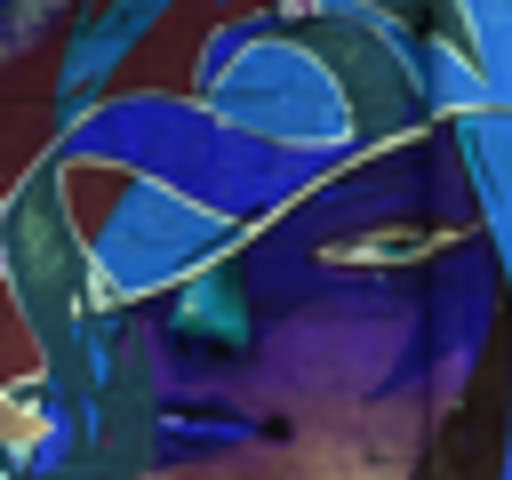}\label{fig:annotation:invalid}}
    \hfil
    \subfloat[Error of Bicubic]{\includegraphics[width=0.23\textwidth]{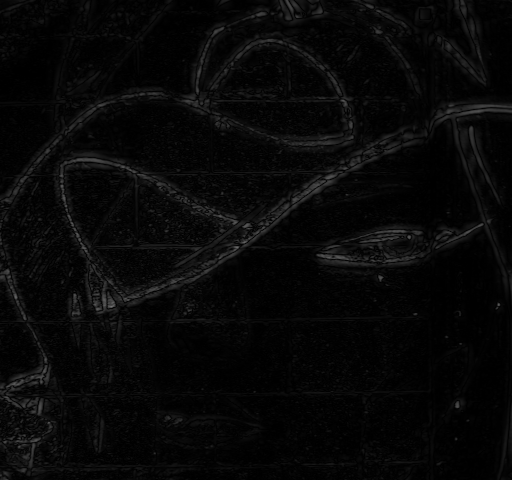}\label{fig:annotation:gt}}
        \hfil 
         \subfloat[GDRRN \cite{HSI:SR:grouped:recursivenet:18}]{\includegraphics[width=0.23\textwidth]{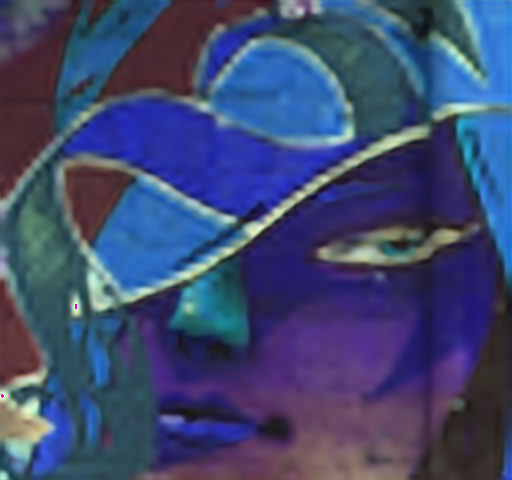}\label{fig:annotation:gt}}
        \hfil 
         \subfloat[Error of GDRRN]{\includegraphics[width=0.23\textwidth]{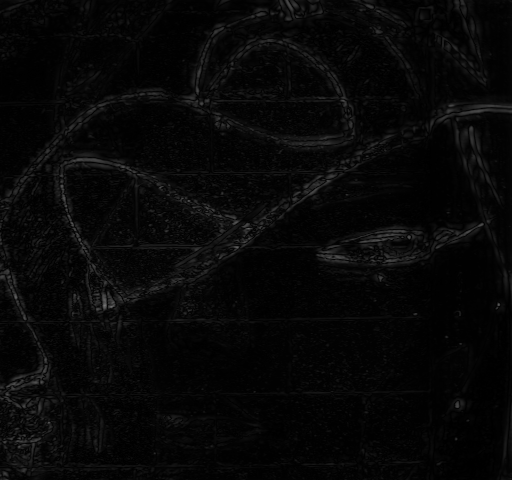}}
    \hfil  \\
           \subfloat[3DFCNN \cite{3d:net:HSI:sr:17}]{\includegraphics[width=0.23\textwidth]{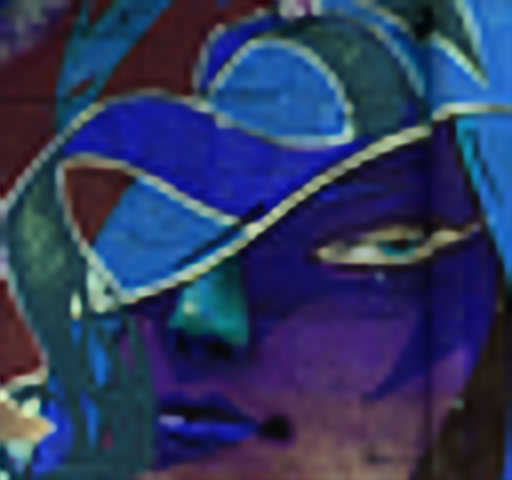}\label{fig:annotation:auxiliary}}
    \hfil
    \subfloat[Error of 3DFCNN]{\includegraphics[width=0.23\textwidth]{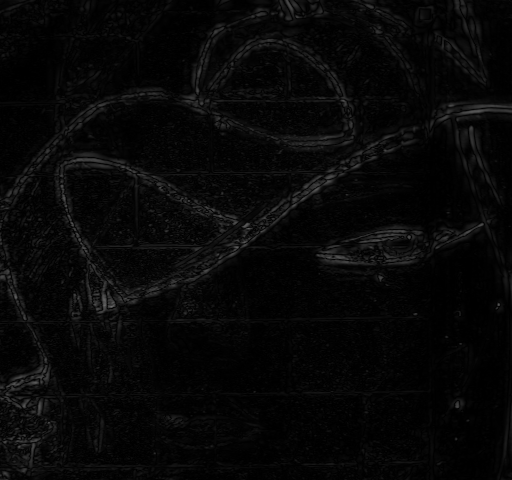}\label{fig:annotation:invalid}}
    \hfil 
    \subfloat[SSPSR \cite{spatial:spectral:prior:20}]{\includegraphics[width=0.23\textwidth]{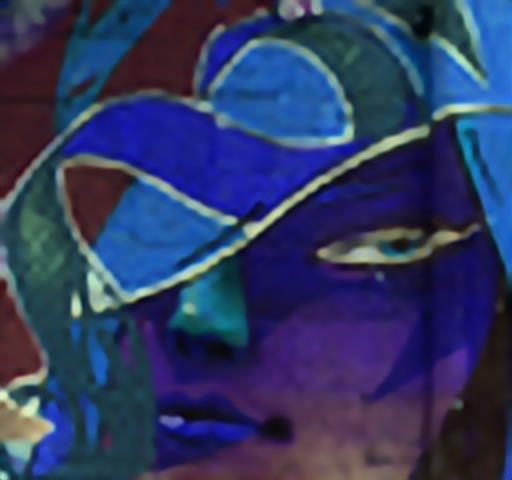}\label{fig:annotation:gt}}
        \hfil
         \subfloat[Error of SSSPSR]{\includegraphics[width=0.23\textwidth]{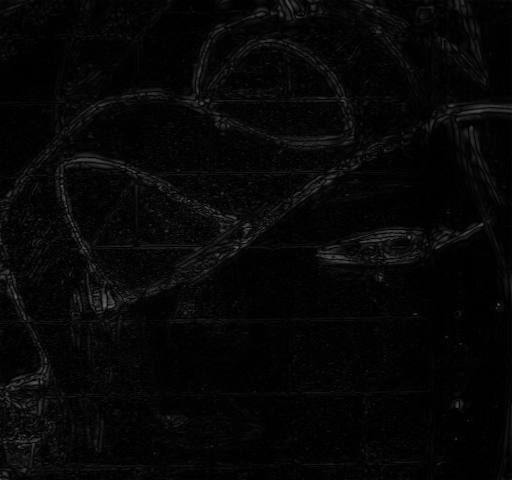}\label{fig:annotation:gt}}
        \hfil \\
    \subfloat[MCNet \cite{2d:3d:net:HSI:SR:20}]{\includegraphics[width=0.23\textwidth]{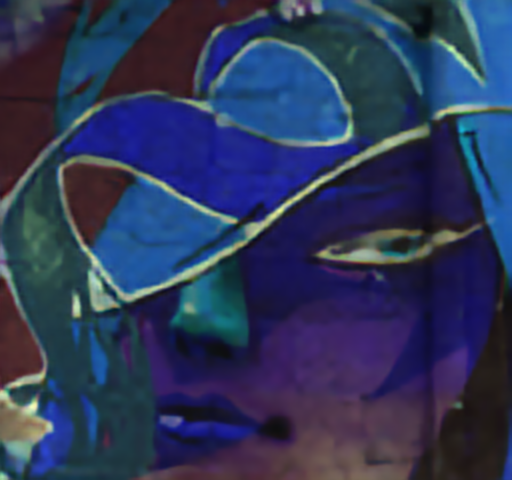}\label{fig:annotation:gt}}
        \hfil
         \subfloat[Error of MCNet]{\includegraphics[width=0.23\textwidth]{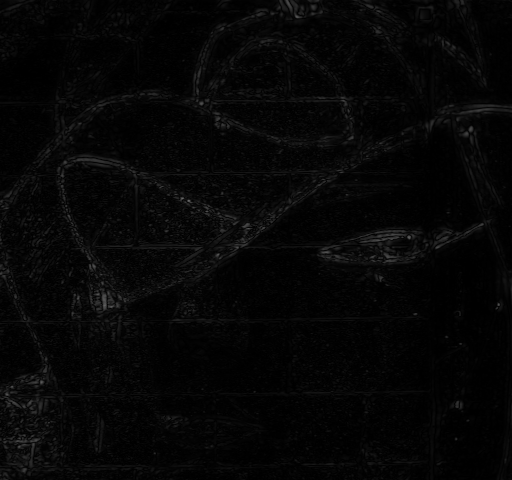}\label{fig:annotation:gt}}
        \hfil 
    \subfloat[Ours]{\includegraphics[width=0.23\textwidth]{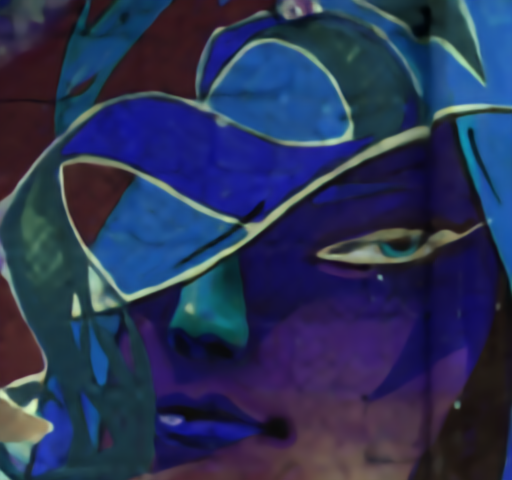}\label{fig:annotation:gt}}
        \hfil
         \subfloat[Error of Ours]{\includegraphics[width=0.23\textwidth]{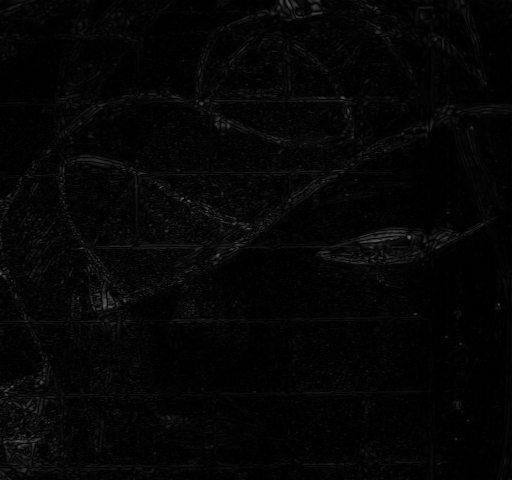}\label{fig:annotation:gt}}
        \hfil 
    \caption{Exemplar results of $\times 8$ by our method and all comparison methods. The error is L2 distance to the ground-truth pixel values, averaged over the three bands. }
    \label{fig:result2}
\end{figure*}

We also provide more visual results on the NTIRE2020 datasets. For visualization, we use the same method as in the main paper. More specifically, we sample the 5th band, the 15th band and the 25th band of the hyperspectral image and assemble them together as an RGB image for visualization. The results of all methods on two different images for scaling factor $\times 8$ are provided in Fig. \ref{fig:result1} and Fig. \ref{fig:result2}. To facilitate the comparison, we also show the error maps of all methods. The values in the error maps are the L2 distance between the predicted pixel values and the ground-truth pixel values, averaged over the three bands. It is clear from the visual results that our method generates better results than other methods. For instance, it produces sharper boundaries and less artefact.

\begin{table*}[!tb]
  \centering
  \begin{tabular}{lcccccccccccccc}
\toprule
& \multicolumn{2} {c} {Components} & \multicolumn{5} {c} {Metrics} \\
Methods & RGB\_SR & \emph{Spec Mixup} & SSL & RMSE $\downarrow$ & CC $\uparrow$ & MPSNR $\uparrow$ & MSSIM $\uparrow$ & ERGAS $\downarrow$ & SAM $\downarrow$\\ \midrule
Ours  & & & &0.01451 & 0.99158 & 40.83762 & 0.95924 & 4.03386 & 4.16312\\ 
Ours  & \cmark & & & 0.01309 & 0.99282 & 41.64471 & 0.96379 & 3.70783 & 4.04210 \\ \cdashline{1-10}
Ours &  &\cmark & & 0.01353 & 0.99237 & 41.51967 & 0.96175 & 3.78138 & 3.75827 \\ 
Ours  &  \cmark & \cmark& &0.01253 & 0.99309 & 42.01961 & 0.96479 & 3.58920  & 3.73269 \\  \cdashline{1-10}
Ours &  \cmark & & \cmark& 0.01281 &  0.99296 & 41.81709  & 0.96483 & 3.64962 & 3.88154\\ 
Ours (final) &  \cmark &\cmark &\cmark & \textbf{0.011909}& \textbf{0.99354} & \textbf{42.35848} & \textbf{0.96679} & \textbf{3.44471} & \textbf{3.78471}\\  \midrule
BicubicInt. & - & - & - & 0.01856 & 0.98682 & 38.73800 & 0.94197 & 5.27190 & 4.17591 \\ 
GDRRN~\cite{HSI:SR:grouped:recursivenet:18}  & - & - & - & 0.02462 & 0.97819 & 36.27752 & 0.90681 &  7.00437 & 8.77122\\ 
3DFCNN~\cite{3d:net:HSI:sr:17}  & - & - & - & 0.01738 &  0.98182 & 38.39284 &  0.94719 & 6.70559  & 6.93475\\ 
SSPSR~\cite{spatial:spectral:prior:20}  & - & - & - & 0.01448 & 0.99152 & 40.91316 & 0.95769 & 4.04064 & 4.07571 \\
MCNet~\cite{2d:3d:net:HSI:SR:20}  & - & - & - & 0.01461 & 0.99046 & 40.73858 & 0.95691 & 4.16595 & 4.59995\\ 
\bottomrule 
\end{tabular} 
\caption{Results of our method and other comparison methods on the CAVE dataset in the semi-sueprvised setting for \textbf{the $\times \mathbf{4}$ case}.} 
  \label{tab:semi:cave}
\end{table*} 

\begin{table*}[!tb]
  \centering
  \begin{tabular}{lcccccccccccccc}
\toprule
& \multicolumn{2} {c} {Components} & \multicolumn{5} {c} {Metrics} \\
Methods & RGB\_SR & \emph{Spec Mixup} & SSL & RMSE $\downarrow$ & CC $\uparrow$ & MPSNR $\uparrow$ & MSSIM $\uparrow$ & ERGAS $\downarrow$ & SAM $\downarrow$\\ \midrule
Ours  & & & & 0.014064 & 0.95902 & 40.46614 & 0.92537  & 3.17093 & 2.55144\\ 
Ours  & \cmark & & & 0.01371 & 0.95985 & 40.70068 & 0.92700 & 3.09505 & 2.54837\\ \cdashline{1-10}
Ours &  &\cmark & &0.01375 &0.95951 & 40.63699 &0.92726  & 3.12395 & 2.54555\\ 
Ours  &  \cmark & \cmark& & 0.01345 & 0.96077 & 40.80595 & 0.92896 & 3.06239 & 2.53140\\  \cdashline{1-10}
Ours &  \cmark & & \cmark& 0.01359 & 0.96003 & 40.75872  & 0.92855  & 3.07786 & 2.54378 \\ 
Ours (final) &  \cmark &\cmark &\cmark & \textbf{0.01331} &\textbf{0.96111} & \textbf{40.93154} & \textbf{0.93004} & \textbf{3.01392} & \textbf{2.52215} \\  \midrule
BicubicInt. & - & - & - & 0.01678 & 0.94994 &38.89758 & 0.90925 &3.80698  & 2.61754 \\
GDRRN~\cite{HSI:SR:grouped:recursivenet:18}  & - & - & - & 0.01609 & 0.94356 & 38.69531 & 0.91265 & 4.30316 & 3.05361 \\ 
3DFCNN~\cite{3d:net:HSI:sr:17}  & - & - & - & 0.01578 & 0.95213 & 39.34413 & 0.91655 & 3.61724 & 2.72764 \\ 
SSPSR~\cite{spatial:spectral:prior:20}  & - & - & - & 0.01427 & 0.95818  & 40.32095 & 0.92356 & 3.22745 & 2.55649 \\
MCNet~\cite{2d:3d:net:HSI:SR:20}  & - & - & - & 0.01468 & 0.95779  & 40.18739  & 0.92140 & 3.26059 & 2.67236 \\
\bottomrule 
\end{tabular} 
\caption{Results of our method and other comparison methods on the Harvard dataset in the semi-supervised setting for \textbf{the $\times \mathbf{4}$ case}.} 
  \label{tab:semi:harvard}
\end{table*} 

\begin{table*}[!tb]
  \centering
  \begin{tabular}{lcccccccccccccc}
\toprule
& \multicolumn{2} {c} {Components} & \multicolumn{5} {c} {Metrics} \\
Methods & RGB\_SR & \emph{Spec Mixup} & SSL & RMSE $\downarrow$ & CC $\uparrow$ & MPSNR $\uparrow$ & MSSIM $\uparrow$ & ERGAS $\downarrow$ & SAM $\downarrow$\\ \midrule
Ours  & & & & 0.01602 & 0.99114  & 38.31541 & 0.93878 & 2.20746 & 1.14009\\ 
Ours  & \cmark & & & 0.01518 & 0.99198 & 38.78697 & 0.94278 & 2.10917  & 1.14217 \\ \cdashline{1-10}
Ours &  &\cmark & & 0.01563 & 0.99156 & 38.58965  & 0.94044 & 2.14927 & 1.14638\\ 
Ours  &  \cmark & \cmark& & 0.01509 & 0.99202 & 38.83281 & 0.94287 & 2.09676 & 1.13495\\  \cdashline{1-10}
Ours &  \cmark & & \cmark& 0.01526 & 0.99194 & 38.72326 & 0.94269  & 2.11944 & 1.13183 \\ 
Ours (final) &  \cmark &\cmark &\cmark &\textbf{0.01485}  &\textbf{0.99226} & \textbf{38.96572} & \textbf{0.94413} & \textbf{2.06742} &  \textbf{1.12279} \\  \midrule
BicubicInt. & - & - & - & 0.02353 & 0.98297& 34.74012& 0.90050 &  3.19014& 3.89655 \\
GDRRN~\cite{HSI:SR:grouped:recursivenet:18}  & - & - & - & 0.01973  & 0.98578 & 36.07932  & 0.91738 & 2.81751 & 2.19765\\ 
3DFCNN~\cite{3d:net:HSI:sr:17}  & - & - & - & 0.02083 & 0.98732 & 35.66309 & 0.91799 & 2.82460 & 1.69877\\ 
SSPSR~\cite{spatial:spectral:prior:20}  & - & - & - & 0.01635 & 0.99078 & 38.07401 & 0.93691  & 2.25392 & 1.43442 \\
MCNet~\cite{2d:3d:net:HSI:SR:20}  & - & - & - & 0.01680 & 0.99051 & 38.02486 & 0.93599  & 2.28342 & 1.45320 \\
\bottomrule 
\end{tabular} 
\caption{Results of our method and other comparison methods on the NTIRE2020 dataset in the semi-supervised setting for \textbf{the $\times \mathbf{4}$ case}.} 
  \label{tab:semi:NTIRE2020}
\end{table*} 

\begin{table*}[!tb]
  \centering
  \begin{tabular}{lcccccccccccccc}
\toprule
& \multicolumn{2} {c} {Components} & \multicolumn{5} {c} {Metrics} \\
Methods & RGB\_SR & \emph{Spec Mixup}  & RMSE $\downarrow$ & CC $\uparrow$ & MPSNR $\uparrow$ & MSSIM $\uparrow$ & ERGAS  $\downarrow$ & SAM $\downarrow$\\ \midrule
Ours  & &  & 0.01196& 0.99350 & 42.38359 & 0.96631 & 3.45903 &4.00592\\ 
Ours  & \cmark &  &0.01109 & 0.99422 &42.73668 & 0.96869 & 3.35884&4.17068  \\ \cdashline{1-10}
Ours &  &\cmark  &0.01134 & 0.99397 & 42.88402 & 0.96798 & 3.28051 & \textbf{3.47102}\\\cdashline{1-10}
Ours (final) &  \cmark &\cmark  &\textbf{0.01068}&\textbf{0.99447}  & \textbf{43.32421} & \textbf{0.96980} & \textbf{3.11799} & 3.68885\\  \midrule
GDRRN~\cite{HSI:SR:grouped:recursivenet:18}  & - & - &  0.01629& 0.98975 & 39.74705 &0.94418  &  4.52683 &5.39660 \\ 
3DFCNN~\cite{3d:net:HSI:sr:17}  & - & - & 0.01583& 0.98538 & 39.21786 & 0.95165 & 5.41798 &6.49516\\ 
SSPSR~\cite{spatial:spectral:prior:20}  & - & - & 0.01245 & 0.99317 & 42.13787 & 0.96457 &3.55146 &3.83398\\
MCNet~\cite{2d:3d:net:HSI:SR:20}  & - & - &  0.01245& 0.99283 &42.25978  & 0.96465 & 3.56246  & 3.84976\\ 
\bottomrule 
\end{tabular} 
\caption{Results of our method and other comparison methods on the CAVE dataest in the fully supervised case for \textbf{the $\times \mathbf{4}$ case}.} 
  \label{tab:fullsup:cave}
\end{table*} 

\begin{table*}[!tb]
  \centering
  \begin{tabular}{lcccccccccccccc}
\toprule
& \multicolumn{2} {c} {Components} & \multicolumn{5} {c} {Metrics} \\
Methods & RGB\_SR & \emph{Spec Mixup} &  RMSE $\downarrow$ & CC $\uparrow$ & MPSNR $\uparrow$ & MSSIM $\uparrow$ & ERGAS  $\downarrow$ & SAM $\downarrow$\\ \midrule
Ours  & & & 0.01344 & 0.96101 & 40.91014 &0.92836 & 3.01039 &2.50704\\ 
Ours  & \cmark  & & 0.01325& 0.96165 &41.03709 &0.92949 &2.96643 &  2.49562\\ \cdashline{1-10}
Ours &  &\cmark  & \textbf{0.01317}& \textbf{0.96200} & \textbf{41.08568}& \textbf{0.93056} & \textbf{2.95718} & \textbf{2.49771}\\ 
Ours (final) &  \cmark &\cmark &0.01321& 0.96178 &41.05925  & 0.93016 & 2.96496 & 2.49897\\  \midrule
GDRRN~\cite{HSI:SR:grouped:recursivenet:18}  & - & - &0.01484  &0.95102  &39.62759 & 0.92013 & 3.67932&2.78624\\ 
3DFCNN~\cite{3d:net:HSI:sr:17}  & - & - & 0.01519 &0.95411 & 39.66271 & 0.91944 &3.47738 &2.63534\\ 
SSPSR~\cite{spatial:spectral:prior:20}  & - & - & 0.01352 & 0.96059 &40.81499  &  0.92806&3.05007&2.51185\\
MCNet~\cite{2d:3d:net:HSI:SR:20}  & - & - &0.01405& 0.96009&  40.59229&0.92658 &3.10529 & 2.59147\\ 
\bottomrule 
\end{tabular} 
\caption{Results of our method and other comparison methods on the Harvard dataset in the fully supervised case for \textbf{the $\times \mathbf{4}$ case}.} 
  \label{tab:fullsup:harvard}
\end{table*} 

\begin{table*}[!tb]
  \centering
  \begin{tabular}{lcccccccccccccc}
\toprule
& \multicolumn{2} {c} {Components} & \multicolumn{5} {c} {Metrics} \\
Methods & RGB\_SR & \emph{Spec Mixup} & SSL & RMSE $\downarrow$ & CC $\uparrow$ & MPSNR $\uparrow$ & MSSIM $\uparrow$ & ERGAS  $\downarrow$ & SAM $\downarrow$\\ \midrule
Ours  & & & & 0.02459 & 0.97233 & 35.89888 & 0.90591 & 7.11644 & 7.50539\\ 
Ours  & \cmark & & & 0.02300 & 0.97545 & 36.50494 & 0.91358  & 6.62181 &  6.89221 \\ \cdashline{1-10}
Ours &  &\cmark & & 0.02339 & 0.97705 &  36.64775 & 0.91078 & 6.45075 & 6.40468\\ 
Ours  &  \cmark & \cmark& & 0.02209 & 0.97923 & 37.14066 & 0.91748 & 6.14668  & 6.24141 \\  \cdashline{1-10}
Ours &  \cmark & & \cmark& 0.02237 & 0.97732 & 36.87625 & 0.91685 & 6.34934 & 6.31945\\ 
Ours (final) &  \cmark &\cmark &\cmark & \textbf{0.02154}& \textbf{0.97963} & \textbf{37.35599} & \textbf{0.92070} & \textbf{6.00466} & \textbf{5.62054}\\  \midrule
BicubicInt. & - & - & - & 0.03042 & 0.96665 & 34.22211 & 0.87386 & 8.43509 & 5.89620 \\ 
GDRRN~\cite{HSI:SR:grouped:recursivenet:18}  & - & - & - & 0.03473 & 0.95693 & 32.93635 & 0.83473 & 9.85545 & 11.05589 \\ 
3DFCNN~\cite{3d:net:HSI:sr:17}  & - & - & - & 0.02920 & 0.90523 & 32.90242 & 0.87942 &  16.72651 & 10.43919\\ 
SSPSR~\cite{spatial:spectral:prior:20}  & - & - & - & 0.02485 & 0.97260 & 35.88966 & 0.89926 & 7.03941 & 7.33639 \\
MCNet~\cite{2d:3d:net:HSI:SR:20}  & - & - & - & 0.02802 & 0.95804 & 34.31165 & 0.88028 & 10.29851 & 7.68969 \\ 
\bottomrule 
\end{tabular} 
\caption{Results of our method and other comparison methods on the CAVE dataset in the semi-sueprvised setting for \textbf{the $\mathbf{\times 8}$ case}.} 
  \label{tab:semi:cave:x8}
\end{table*} 

\begin{table*}[!tb]
  \centering
  \begin{tabular}{lcccccccccccccc}
\toprule
& \multicolumn{2} {c} {Components} & \multicolumn{5} {c} {Metrics} \\
Methods & RGB\_SR & \emph{Spec Mixup} & SSL & RMSE $\downarrow$ & CC $\uparrow$ & MPSNR $\uparrow$ & MSSIM $\uparrow$ & ERGAS $\downarrow$ & SAM $\downarrow$\\ \midrule
Ours  & & & & 0.02263 & 0.92668 & 36.65405 & 0.86476 & 4.85976 & 2.91037\\ 
Ours  & \cmark & & &0.02127 & 0.93172 &  37.14154 & 0.87229 & 4.58089  & 2.88299 \\ \cdashline{1-10}
Ours &  &\cmark & &0.02219 &0.92789 & 36.85938 & 0.86725  & 4.76278 & 2.89283\\ 
Ours  &  \cmark & \cmark& &0.02117 & 0.93241 & 37.18768 & 0.87294 & 4.55818 &\textbf{2.85549}\\  \cdashline{1-10}
Ours &  \cmark & & \cmark& 0.02116 & 0.93218 & 37.21642  & 0.87269 & 4.55045 & 2.88124\\ 
Ours (final) &  \cmark &\cmark &\cmark & \textbf{0.02100} &\textbf{0.93450} & \textbf{37.35231} & \textbf{0.87345} & \textbf{4.54990} & 2.88230 \\  \midrule
BicubicInt. & - & - & - & 0.02495  & 0.91446 & 35.74092 & 0.84966 & 5.47721 & 3.00931 \\
GDRRN~\cite{HSI:SR:grouped:recursivenet:18}  & - & - & - & 0.02389 & 0.91131 & 35.64417 & 0.85316 & 5.72871  & 3.41440 \\ 
3DFCNN~\cite{3d:net:HSI:sr:17}  & - & - & - & 0.02379 & 0.91813 & 36.05515 & 0.85651 & 5.21924 & 3.09288 \\ 
SSPSR~\cite{spatial:spectral:prior:20}  & - & - & - & 0.02282 & 0.92266 & 36.45639 & 0.86174 & 4.99781 & 3.07039 \\
MCNet~\cite{2d:3d:net:HSI:SR:20}  & - & - & - & 0.02348 & 0.92343 & 36.39212 & 0.85940 & 5.05722 & 3.14259 \\
\bottomrule 
\end{tabular} 
\caption{Results of our method and other comparison methods on the Harvard dataset in the semi-supervised setting for \textbf{the $\mathbf{\times 8}$ case}.} 
  \label{tab:semi:harvard:x8}
\end{table*} 

\begin{table*}[!tb]
  \centering
  \begin{tabular}{lcccccccccccccc}
\toprule
& \multicolumn{2} {c} {Components} & \multicolumn{5} {c} {Metrics} \\
Methods & RGB\_SR & \emph{Spec Mixup} & SSL & RMSE $\downarrow$ & CC $\uparrow$ & MPSNR $\uparrow$ & MSSIM $\uparrow$ & ERGAS $\downarrow$ & SAM $\downarrow$\\ \midrule
Ours  & & & &0.02969 & 0.97146 & 32.83078 & 0.84733 &  4.04429& 1.88851\\ 
Ours  & \cmark & & &0.02839& 0.97343 & 33.23283& 0.85401 & 3.89444 & 1.83394\\ \cdashline{1-10}
Ours &  &\cmark & &0.03004 & 0.97055 &  32.75617 & 0.84575 & 4.08821 & 1.90378\\ 
Ours  &  \cmark & &\cmark & 0.02800 & 0.97355 & 33.23400  & 0.85870 & 3.88999 & 1.83009\\  \cdashline{1-10}
Ours &  \cmark &\cmark & &0.02809 & 0.97385 &33.33009 & 0.85599 & 3.85485& 1.79814\\ 
Ours (final) &  \cmark &\cmark &\cmark &\textbf{0.02799}  &\textbf{0.97560} & \textbf{33.45620} & \textbf{0.86001} & \textbf{3.84456} &  \textbf{1.76943} \\  \midrule
BicubicInt. & - & - & - &0.03961  & 0.95195& 29.95891&0.78926  & 5.45947 & 5.24567 \\
GDRRN~\cite{HSI:SR:grouped:recursivenet:18}  & - & - & - &  0.03596&  0.95545& 30.67236 &0.80217 &5.12654 &3.21549 \\ 
3DFCNN~\cite{3d:net:HSI:sr:17}  & - & - & - &0.38575 &-0.57977 &9.17534 & -0.47680 &61.62476 & 163.56878\\ 
SSPSR~\cite{spatial:spectral:prior:20}  & - & - & - &0.03268 &0.96529  &  31.78969 & 0.82646 & 4.49521  & 2.25976 \\
MCNet~\cite{2d:3d:net:HSI:SR:20}  & - & - & - &0.03276 & 0.96623 & 31.96291 & 0.82932  & 4.41695 & 2.38619 \\
\bottomrule 
\end{tabular} 
\caption{Results of our method and other comparison methods on the NTIRE2020 dataset in the semi-supervised setting for \textbf{the $\mathbf{\times 8}$ case}.} 
  \label{tab:semi:NTIRE2020:x8}
\end{table*} 


\end{document}